\documentclass[10pt,twocolumn,letterpaper]{article}

\usepackage{cvpr}              

\usepackage{graphicx}
\usepackage{amsmath}
\usepackage{amssymb}
\usepackage{booktabs}
\usepackage{bbm}
\usepackage{multirow}
\usepackage{pifont}
\usepackage{amssymb}
\usepackage{overpic}
\newcommand{\cmark}{\ding{51}}%
\newcommand{\xmark}{\ding{55}}%

\usepackage[pagebackref,breaklinks,colorlinks]{hyperref}

\usepackage[capitalize]{cleveref}
\crefname{section}{Sec.}{Secs.}
\Crefname{section}{Section}{Sections}
\Crefname{table}{Table}{Tables}
\crefname{table}{Tab.}{Tabs.}
 \newcommand{\Hquad}{\hspace{0.2em}}

\def\cvprPaperID{5752} 
\def\confName{ICCV}
\def\confYear{2023}
\begin{document}

\title{LDL: Line Distance Functions for Panoramic Localization}
\author{Junho Kim\textsuperscript{1}, Changwoon Choi\textsuperscript{1}, Hojun Jang\textsuperscript{1}, and Young Min Kim\textsuperscript{1, 2}
\and {\small \phantom{ }} \vspace{-1em}\\
\textsuperscript{1} {\small Dept. of Electrical and Computer Engineering, Seoul National University} \\
\textsuperscript{2} {\small Interdisciplinary Program in Artificial Intelligence and INMC, Seoul National University} \\
{\tt\small 82magnolia@snu.ac.kr, changwoon.choi00@gmail.com, \{j12040208, youngmin.kim\}@snu.ac.kr}
}
\maketitle

\begin{abstract}
We introduce LDL, a fast and robust algorithm that localizes a panorama to a 3D map using line segments.
LDL focuses on the sparse structural information of lines in the scene, which is robust to illumination changes and can potentially enable efficient computation.
While previous line-based localization approaches tend to sacrifice accuracy or computation time, our method effectively observes the holistic distribution of lines within panoramic images and 3D maps.
Specifically, LDL matches the distribution of lines with 2D and 3D line distance functions, which are further decomposed along principal directions of lines to increase the expressiveness.
The distance functions provide coarse pose estimates by comparing the distributional information, where the poses are further optimized using conventional local feature matching.
As our pipeline solely leverages line geometry and local features, it does not require costly additional training of line-specific features or correspondence matching.
Nevertheless, our method demonstrates robust performance on challenging scenarios including object layout changes, illumination shifts, and large-scale scenes, while exhibiting fast pose search terminating within a matter of milliseconds.
We thus expect our method to serve as a practical solution for line-based localization, and complement the well-established point-based paradigm.
The code for LDL is available through the following link: \url{https://github.com/82magnolia/panoramic-localization}.

\end{abstract}

\section{Introduction}
\begin{figure}[t]
  \centering
    \includegraphics[width=\linewidth]{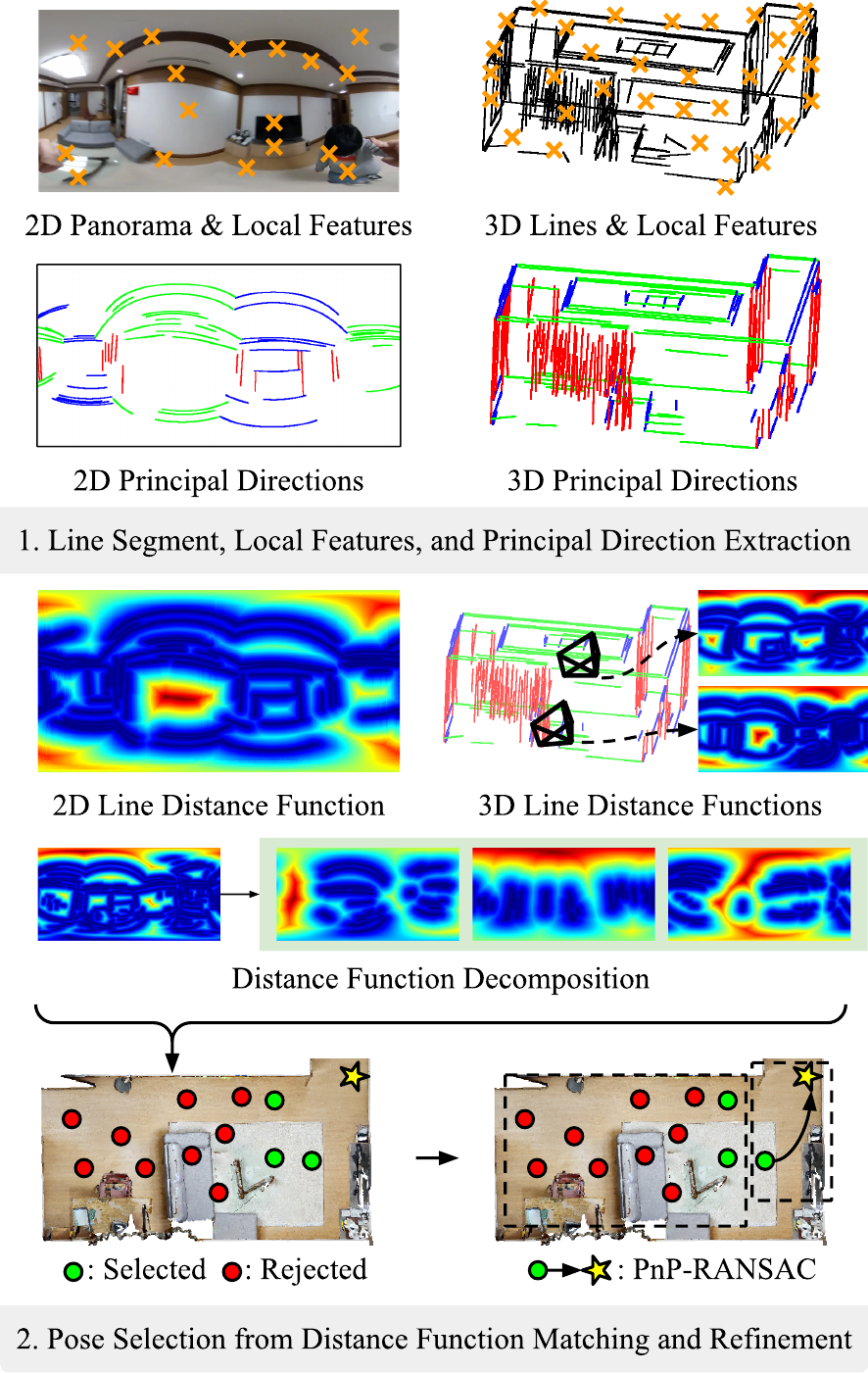}
   \caption{Overview of our approach. LDL assumes a 3D map equipped with lines and local features, and similarly preprocesses the 2D panorama prior to localization. LDL then selects candidate poses by matching 2D, 3D line distance functions through decomposition along principal directions that effectively represent the sparse geometry of lines. Finally, the selected poses are refined via local feature matching~\cite{sarlin2020superglue} and PnP-RANSAC~\cite{ransac, epnp}. }
   \label{fig:overview}
\vspace{-1em}
\end{figure}

Estimating the location of a mobile device or agent with respect to a 3D map, widely referred to as visual localization, has vast applications in robotics and AR/VR.
Compared to perspective images, which are more widely used for localization, panorama images provide a $360^\circ$ field of view that contains ample visual evidence from the holistic scene context. 
In this light, there have been recent advances in visual localization using panoramic images~\cite{gosma, gopac, piccolo, cpo} that demonstrate reasonably stable localization, with state-of-the-art methods leveraging a two-step process of candidate pose selection and refinement~\cite{cpo,sarlin2019coarse}.
Nevertheless, many existing methods for this task have limitations in computational efficiency and robustness, mainly stemming from the costly or unstable pose selection process.
As global feature descriptors~\cite{robust_retrieval,netvlad} or a large number of colored points~\cite{piccolo, cpo} are the main components for this step, the pipelines can be memory and compute intensive or fragile to large illumination changes~\cite{cpo,piccolo}.

To overcome such limitations, we explore the alternative direction of using \textit{lines} as the major cue for panoramic localization.
Lines have a number of desirable properties compared to commonly used raw color, semantic labels or learned global features~\cite{sarlin2019coarse,gosma,piccolo}.
First, due to the long-standing work in line segment extraction~\cite{LSD, lineseg_1, lineseg_2, lineseg_3}, it is cheap and stable to extract line segments even amidst dramatic changes in illumination or moderate motion blur.
Second, lines are sparse representations of a scene and can potentially lead to small memory consumption and computation.
Nevertheless, line segments alone are visually ambiguous compared to other localization cues (color, global features, etc.), which makes them harder to tailor for successful localization.
While there exist prior works in line-based visual localization~\cite{line_transformer,line_chamfer,line_refinement}, many focus on using lines for \textit{pose refinement} after finding coarse poses from conventional global feature comparisons~\cite{line_transformer,line_refinement} or exhibit unstable performance compared to conventional point-based methods~\cite{line_chamfer}.
Further, prior works often involve expensive line-specific feature extraction to distinguish contexts and establish one-to-one line correspondences~\cite{line_transformer}.

LDL is a fast and robust localization method that leverages the holistic context from lines in panoramas and 3D maps to effectively find the camera pose.
In contrast to previous works~\cite{line_refinement,line_transformer}, we retain our focus on using line segments for \textit{pose search} based on the observation that conventional point-based matching~\cite{superpoint,sarlin2020superglue} performs stably once given a good initial pose. 
As shown in Figure~\ref{fig:overview}, given a panoramic image of an unknown location, we utilize the distribution of extracted line segments and compare it against those in the pre-captured 3D map.
First, the candidate pose selection step rapidly evaluates an immense set of poses within a matter of milliseconds and selects the coarse poses to further optimize.
Here LDL compares the distribution of lines in 2D and 3D evaluated on their spherical projections using distance functions, as shown in Figure~\ref{fig:overview}.
The distance function imbues relative spatial context even in featureless regions and quickly matches poses without establishing explicit correspondences between detected lines.
We further enhance the discriminative power of distance functions by decomposition, and separately evaluate lines aligned with each principal directions.
Once a small set of initial poses are found, LDL refines them with PnP-RANSAC~\cite{ransac, epnp}, where we leverage powerful local features from recent works~\cite{sarlin2020superglue,superpoint} to establish good 2D-3D correspondences.

We evaluate LDL in various indoor scenes where it performs competitively against all tested baselines while demonstrating robust performance in scenes with object changes or large illumination shifts.
Further, LDL exhibits an order-of-magnitude faster runtime compared to global feature comparison~\cite{netvlad,robust_retrieval,openibl} due to the efficient formulation.
By only using the geometric information of lines and pre-trained visual features, we expect LDL to serve as a practical localization algorithm that could enhance and complement existing visual localization techniques.

\section{Related Work}


\paragraph{Line-Based Localization}
Inspired by abundant straight-lines and rectangular structures in man-made objects, many works attempt visual localization with line segments~\cite{line_chamfer, line_transformer, sem_line, taubner2020lcd, l2d2, line_refinement}.
Micusik et al.~\cite{line_chamfer} utilize the line segments extracted from the 3D model to directly match line segments in images by comparing the Chamfer distance in 2D and 3D.
However, lines, even when perfectly matched, are inherently subject to ambiguity along the line direction.
Yoon et al.~\cite{line_transformer} suggest removing such ambiguities by treating points on a line segment as verbal tokens in natural language processing, where line features are learned using  Transformers~\cite{transformer}.
Such learning-based approaches are trained with a database of pose-annotated images or require additional computation~\cite{line_transformer, sem_line,line_refinement}.
Further, these approaches only use lines for pose refinement, assuming a coarse pose estimate to be given via global feature comparisons~\cite{netvlad,openibl}.
LDL takes a different approach and focuses on robust \textit{pose selection} based on lines.
We compare LDL against existing approaches
for line-based localization, where LDL performs competitively against these
methods while balancing robustness and efficiency.

\vspace{-1em}
\paragraph{Point-Based Localization}
Most visual localization algorithms follow a point-based paradigm, focusing on sparse feature point correspondences~\cite{hierarchical_scene, active_search_eccv, active_search, learned_loc_1, learned_loc_2, learned_loc_3, learned_loc_4, loc_feat_1, loc_feat_2, sarlin2019coarse, score}, dense matching via coordinate regression of scene points~\cite{hierarchical_scene, dsac}, or minimizing color discrepancies of dense 3D points via gradient descent ~\cite{piccolo, cpo}.
Conventional approaches using a perspective camera input take a two-step approach, where coarse poses are first estimated using global feature descriptors~\cite{openibl, netvlad} and refined with PnP-RANAC from local feature matches~\cite{sift, sarlin2020superglue, superpoint} or dense matches from scene coordinate regression~\cite{hierarchical_scene, score}.
Recent panoramic localization methods~\cite{piccolo, cpo, gosma, gopac} also follow a similar two-step approach, where exemplary methods find candidate poses via color distribution matching and refine them using gradient descent optimization~\cite{piccolo, cpo}.
While these algorithms can robustly handle a modest range of scene changes due to the holistic view from panoramas, the algorithms can still fail with significant changes in illumination.
We compare LDL against exemplary point-based methods and demonstrate that line segments could be effectively utilized for accurate and robust localization even without the costly calculation of global features or color matching.

\section{Method}
\begin{figure}[t]
  \centering
    \includegraphics[width=\linewidth]{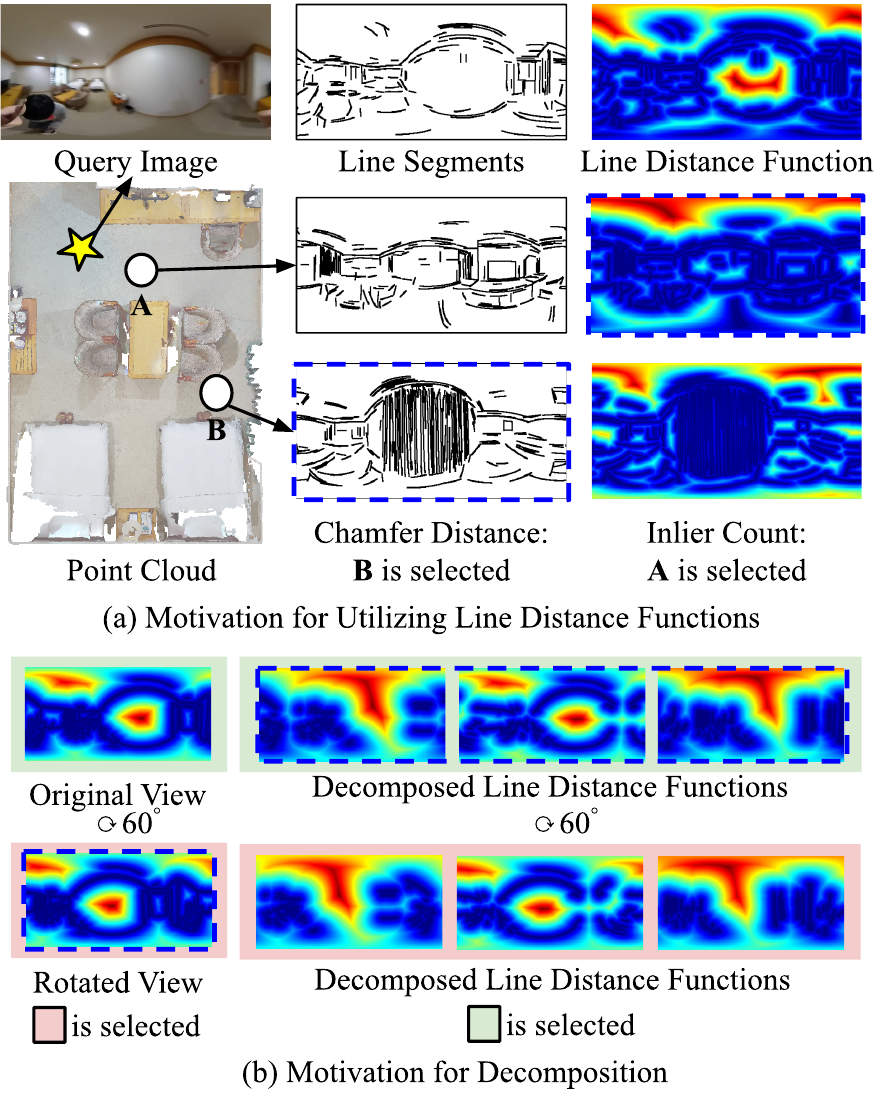}
   \caption{Motivation for (a) utilizing and (b) decomposing line distance functions. (a) Line distance functions disambiguate regions with dense lines. Given two candidate poses close (\textbf{A}) and far (\textbf{B}) from ground truth, Chamfer distance falsely favors \textbf{B} near dense lines, whereas distance functions correctly rank the poses. (b) Decomposition further reduces ambiguities from rotation by separately considering line segments with varying directions. Given an original view close to the ground truth (green) and a rotated view (red), the decomposition better distinguishes the two views by correctly selecting the original view over the rotated view.}
   \label{fig:motivation}
   \vspace{-1em}
\end{figure}

LDL aims at finding the pose at which the query image $I$ is taken with respect to a 3D scene, where Figure~\ref{fig:overview} depicts the localization steps taken by LDL.
We first represent the 3D scene using a line map equipped with local feature descriptors for keypoint locations, and similarly acquire line segments and local descriptors for the query image prior to localization (Section~\ref{sec:input prep}).
We then estimate the three principal directions for 2D and 3D by voting, from which we can deduce a set of rotations considering the sign and permutation ambiguity (Section~\ref{sec:rotation}).
Given the fixed set of candidate rotations, we construct an initial set of possible poses incorporating translations. 
We generate the decomposed line distance functions at each pose and choose the promising poses by comparing the distance functions with a robust loss function (Section~\ref{sec:distance}).
As the final step, the selected poses are refined by performing PnP-RANSAC~\cite{ransac} using feature matches~\cite{sarlin2020superglue} with the query image (Section~\ref{sec:refinement}).

\subsection{Localization Input Preparation}
\label{sec:input prep}

\paragraph{Map Building}
LDL operates using a 3D map consisting of line segments and local features.
We build such a map starting from a colored point cloud $P=\{X{,}C\}$.
To obtain the 3D line segments we use the line extraction method from Xiaohu et al.~\cite{3d_lineseg}, which can quickly process point clouds containing millions of points within a few seconds.
We further remove short, noisy line segments from the raw detection with a simple filtering step: given the point cloud bounding box of size $b_x \times b_y \times b_z$, we filter out 3D line segments shorter than $\lambda (b_x + b_y + b_z)/3$ with $\lambda=0.1$ in all our experiments.
The 2D line segments are then filtered with an adaptive length threshold to match the filtering rate of 3D line segments.
Specifically, we choose the threshold value such that the ratio of lines filtered in 2D equals that in 3D.

To obtain local features embedded in the 3D map, we first render synthetic views at various locations using the point cloud color values.
Specifically, we project the input point cloud $P{=}\{X,C\}$ at a virtual camera and assign the measured color $Y(u,v){=}C_n$  at the projected location of the corresponding 3D coordinate $(u,v){=}\Pi(R X_n + t)$ to create the synthetic view $Y$.
We then extract local features for each synthetic view $Y$ using SuperPoint~\cite{superpoint}, and back-project the local features to their 3D locations, which in turn results in keypoint descriptors embedded in 3D space.
Note that while we illustrate map building using a colored point cloud, our setup can also work with line-based SfM maps~\cite{line_sfm_1,line_sfm_2,line_sfm_3} since the input to our pipeline is lines and associated local features.

\paragraph{Panorama Pre-processing}
Similar to map building, we extract line segments and local features from the query panorama image.
We use LSD~\cite{LSD} to acquire line segments, which is a robust line detection algorithm that can stably extract lines even under motion blur or lighting changes.
To remove noisy line detections as in the 3D case, we filter 2D line segments with an adaptive length threshold to match the filtering rate of 3D line segments.
Specifically, for each scene we choose the threshold value such that the ratio of lines filtered in 2D equals that in 3D.
Then, we extract local feature descriptors using SuperPoint~\cite{superpoint}, where the results will later be used for pose refinement in Section~\ref{sec:refinement}.

\subsection{Candidate Rotation Estimation}
\label{sec:rotation}
Given the detected line segments, LDL first estimates a set of feasible rotations by extracting principal directions, which we define as the most common line directions in 2D and 3D.
Let $L_{2D}=\{l\}$ denote the line segments in 2D, where $l=(s, e)$ is a tuple of start point $s \in \mathbb{S}^2$ and end point $e \in \mathbb{S}^2$.
Note that we operate on the spherical projection space and treat lines and points on panoramas as arcs and points on the unit sphere $\mathbb{S}^2$ respectively.
Similarly, let $L_{3D}=\{\tilde{l}\}$ denote the line segments in 3D, with $\tilde{l}=(\tilde{s}, \tilde{e})$ being a tuple containing start and end points $\tilde{s}, \tilde{e} \in \mathbb{R}^3$.

LDL estimates the vanishing point and votes for the principal directions in 2D and 3D.
In 2D we first extract vanishing points by finding the points of intersection of extended 2D line segments.
The 2D principal directions $P_{2D}{=}\{p\}$ are defined as the top $k_{2D}$ vanishing points containing the most incident lines, where $p \in \mathbb{R}^3$ is a unit norm vector denoting the vanishing point location in the sphere.
Similarly, the 3D principal directions $P_{3D}{=}\{\tilde{p}\}$ are defined as the top $k_{3D}$ most common line directions from 3D line segments obtained via voting.
Note that the 3D direction $\tilde{p} \in \mathbb{R}^3$ is also normalized.

LDL estimates the feasible candidate rotations up to uncertainty in the combinatorial ambiguities when matching the principal directions in 2D and 3D.
Specifically, we select triplets of directions from $P_{2D}$ and $P_{3D}$, yielding a total of ${k_{2D}\choose 3} \times {k_{3D} \choose 3} \times {3!} \times {2^3}$ possible combinations, additionally considering the sign and permutation ambiguity.
For each pair of triplets, we apply the Kabsch algorithm~\cite{kabsch} to find the optimal rotation that aligns the 2D directions to 3D directions.
Discarding infeasible rotations that have large mean squared error, we obtain $N_r$ rotations.
The possible rotations are further filtered using line distance function presented in the next section.

\subsection{Line Distance Functions for Pose Selection}
\label{sec:distance}
We propose line distance functions to efficiently evaluate a large pool of poses and select promising candidate poses.
The initial pool of poses is the combination of possible translations with the rotations found in the previous section. 
To this end, $N_t$ translations are chosen within grid partitions of the 3D point cloud, where details are explained in the supplementary material.
The resulting $N_t \times N_r$ poses are ranked using line distance functions.

\paragraph{Distance Function Definition} 
Distance functions are designed to compare the holistic spatial context captured from the large field of view in panorama images.
They are defined for every point including void regions without any lines and can quickly rank poses.
Compared to Chamfer distance or learned line embeddings used in prior work~\cite{line_chamfer,line_transformer}, 
LDL does not attempt pairwise matching between lines, which is often costly and can incur failure modes.
For example, it is ambiguous to correctly match between densely packed lines as shown in Figure~\ref{fig:motivation}\textcolor{red}{a}.

A line distance function is a dense field of distance values to detect lines in the 2D query image or the spherical projection at an arbitrary pose in 3D.
For a point $x$ on the unit sphere $\mathbb{S}^2$, the 2D line distance function is given as

\begin{equation}
	f_{2D}(x ; L_{2D}) = \min_{l \in L_{2D}} D(x, l).
\end{equation}

Here $D(x, l)$ is the spherical distance from $x$ to line segment $l=(s, e)$, namely
\begin{equation}
	D(x, l) = 
\left\{
	\begin{array}{lr}
		\sin^{-1} |\langle x, \dfrac{s \times e}{\|s \times e\|}\rangle|  &  \!\!\!\!\! \mbox{if $x \in \mathcal{Q}(s,e)$}\\
		\min (\cos^{-1} \langle x, e\rangle, \cos^{-1} \langle x, s\rangle) &  \!\!\!\!\! \mbox{otherwise,}
	\end{array}
\right.
\label{eq:line_dist}
\end{equation}
where $\mathcal{Q}(s,e)$ is the spherical quadrilateral formed from $\{s, e, \pm (s\times e) / \|s\times e\|\}$.

\begin{figure}[t]
  \centering
    \includegraphics[width=.85\linewidth]{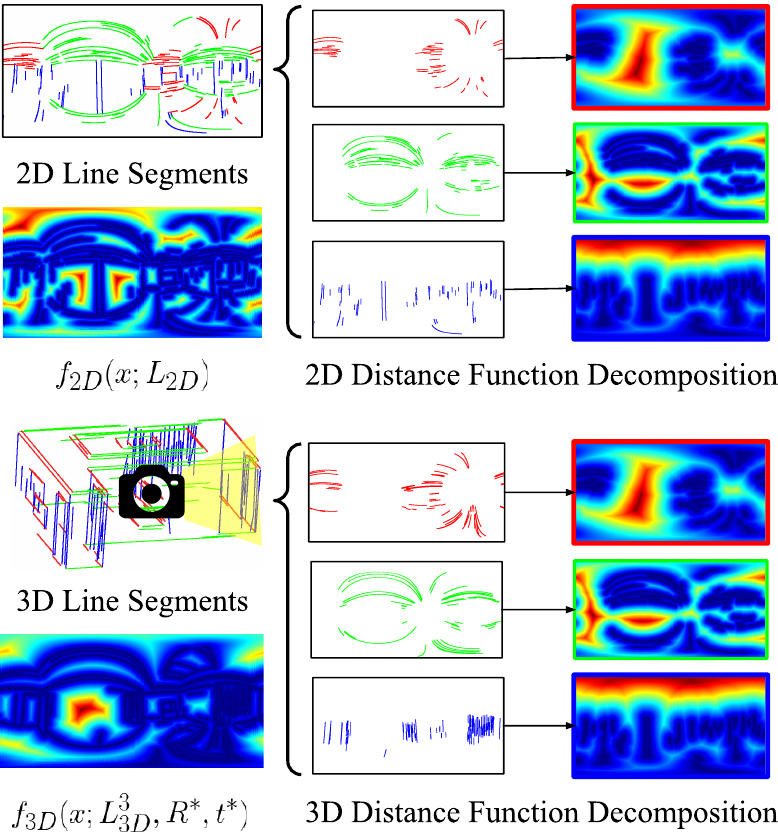}
   \caption{Line distance function visualization and decomposition at the ground truth pose $R^*,t^*$. LDL decomposes distance functions using principal directions and enhances their expressiveness.
   }
   \label{fig:decomp}
\vspace{-1em}
\end{figure}

Similarly, the 3D line distance function is defined for each candidate rotation $R \in SO(3)$ and translation $t \in \mathbb{R}^3$.
Using the spherical projection function $\Pi(\cdot): \mathbb{R}^{3}\rightarrow \mathbb{S}^{2}$ that maps a point in 3D to a point on the unit sphere, the 3D line segment $\tilde{l} = (\tilde{s}, \tilde{e})$ is projected to 2D under the candidate transformation as $l = (\Pi(R\tilde{s} + t), \Pi(R\tilde{e} + t))$.
For simplicity, let $\Pi_L (\tilde{l}; R, t)$ denote the projection of a line segment in 3D to the spherical surface. 
Then the 3D line distance function is defined as follows,
\begin{equation}
	f_{3D}(x; L_{3D}, R, t) = \min_{\tilde{l} \in L_{3D}} D(x, \Pi_L(\tilde{l}; R, t)).
\end{equation}
As shown in Figure~\ref{fig:decomp}, one can expect poses closer to the ground truth to have similar 2D and 3D line distance functions.
Therefore, we evaluate $N_t \times N_r$ poses according to the similarity of line distance functions.

We apply a robust loss function that measures inlier counts to quantify the affinity of the line distance functions.
For each candidate pose $\{R, t\}$ we count the number of points whose distance function differs below a threshold $\tau$,
\begin{equation}
L(R, t) = \! - \!\! \sum_{q \in Q} \mathbbm{1}\{|f_{2D}(q; L_{2D}) - f_{3D}(q; L_{3D}, R, t)| < \tau\},
\label{eq:loss}
\end{equation}
where $\mathbbm{1}\{\cdot\}$ is the indicator function and $Q \subset \mathbb{S}^2$ is a set of query points uniformly sampled from a sphere.
The loss function only considers inlier counts, and thus is robust to outliers from scene changes or line misdetections.
We validate the efficacy of the robust loss function in Section~\ref{sec:perf}.

\paragraph{Distance Function Decomposition}
To further enhance pose search using line distance functions, we propose to decompose the distance functions along three principal directions.
While line distance functions provide useful evidence for line-based localization, they lack a sense of direction as in Figure~\ref{fig:motivation}\textcolor{red}{b}, where the distance functions alone cannot effectively distinguish rotated views at a fixed translation.

We split line segments along the principal directions used for rotation estimation and define separate line distance functions for each group of lines, as shown in Figure~\ref{fig:decomp}.
Recall from Section~\ref{sec:rotation} that each candidate rotation $R$ is obtained from a pair of triplets in 2D and 3D principal directions denoted as $\hat{P}_{2D}^R{=}\{p_1, p_2, p_3\}$ and $\hat{P}_{3D}^R{=}\{\tilde{p}_1, \tilde{p}_2, \tilde{p}_3\}$.
We associate line segments that are parallel to directions in $\hat{P}_{2D}^R, \hat{P}_{3D}^R$, leading to three groups of line segments $L_{2D}^R{=}\{L_{2D}^1, L_{2D}^2, L_{2D}^3\}$ and  $L_{3D}^R{=}\{L_{3D}^1, L_{3D}^2, L_{3D}^3\}$ in 2D and 3D, respectively.
We separately define line distance functions for the three groups using Equation~\ref{eq:line_dist}, namely $f_{2D}(x; L_{2D}^i)$ and $f_{3D}(x; L_{3D}^i, R, t)$ for $i = 1, 2, 3$.
Then the robust loss function in Equation~\ref{eq:loss} can be modified to accommodate the decomposed distance functions,
\vspace{-1em}
\begin{equation}
L(R, \! t) \! = \! - \!\!\sum_{i=1}^3\!\sum_{q \in Q} \!\mathbbm{1}\{|f_{2D}(q; \! L_{2D}^i) - f_{3D}(q;\! L_{3D}^i,\! R,\! t)| \!< \!\tau\}.
\label{eq:decomp}
\end{equation}
We validate the importance of distance function decomposition in Section~\ref{sec:perf}.

\subsection{Candidate Pose Refinement}
\label{sec:refinement}
After we select the top $K$ poses from the pool of $N_t \times N_r$ poses with the loss function values from Equation~\ref{eq:decomp}, we refine them using local feature matching as shown in Figure~\ref{fig:overview}.
Here we utilize the cached local features from Section~\ref{sec:input prep}.
Specifically, for each selected pose we first retrieve the set of visible 3D keypoints at that pose and perform local feature matching against the 2D keypoints in the query image.
In this process we use SuperGlue~\cite{sarlin2020superglue} for feature matching and select the candidate pose with the most matches.
Finally, we apply PnP-RANSAC~\cite{ransac, epnp, pnp_1} on the matched 2D and 3D keypoint coordinates to obtain a refined pose estimate.
Backed by local feature matching that stably operates given decent coarse estimates from line distance functions, LDL can robustly function as an effective localization method which we further verify in Section~\ref{sec:exp}.

\begin{table*}[t]
    \centering
    \resizebox{0.85\linewidth}{!}{
    \begin{tabular}{l|cccccc|cccccc|cccccc}
        \toprule
        {} & \multicolumn{6}{c|}{$t$-error (m)} & \multicolumn{6}{c|}{$R$-error ($^\circ$)} & \multicolumn{6}{c}{Accuracy}\\
        Dataset & PC & SB & CD & LT & CPO & LDL & PC & SB & CD & LT & CPO & LDL & PC & SB & CD & LT & CPO & LDL \\
        \midrule
        Area 1 & 0.02 & 0.02 & 0.12 & 0.02 & \textbf{0.01} & 0.02 & 0.46 & 0.62 & 1.14 & 0.62 & \textbf{0.25} & 0.54 & 0.66 & 0.89 & 0.50 & \textbf{0.90} & \textbf{0.90} & 0.86 \\
        Area 2 & 0.76 & 0.04 & 1.16 & 0.04 & \textbf{0.01} & 0.02 & 2.25 & 0.72 & 11.54 & 0.72 & \textbf{0.27} & 0.66 & 0.45 & 0.76 & 0.35 & 0.74 & \textbf{0.81} & 0.77 \\
        Area 3 & 0.02 & 0.03 & 0.79 & 0.02 & \textbf{0.01} & 0.02 & 0.49 & 0.57 & 4.54 & 0.55 & \textbf{0.24} & 0.54 & 0.57 & \textbf{0.92} & 0.36 & 0.88 & 0.78 & 0.89 \\
        Area 4 & 0.18 & 0.02 & 0.33 & 0.02 & \textbf{0.01} & 0.02 & 4.17 & 0.57 & 1.97 & 0.56 & \textbf{0.28} & 0.48 & 0.49 & \textbf{0.91} & 0.46 & \textbf{0.91} & 0.83 & 0.88 \\
        Area 5 & 0.50 & 0.03 & 0.95 & 0.03 & \textbf{0.01} & 0.02 & 14.64 & 0.69 & 41.84 & 0.65 & \textbf{0.27} & 0.54 & 0.44 & 0.80 & 0.36 & 0.79 & 0.74 & 0.81 \\
        Area 6 & 0.01 & 0.02 & 0.50 & 0.02 & \textbf{0.01} & 0.02 & 0.31 & 0.63 & 1.20 & 0.60 & \textbf{0.18} & 0.50 & 0.69 & \textbf{0.88} & 0.47 & 0.87 & 0.90 & 0.83 \\ \midrule
        Total & 0.03 & 0.03 & 0.73 & 0.02 & \textbf{0.01} & 0.02 & 0.63 & 0.63 & 2.30 & 0.63 & \textbf{0.24} & 0.53 & 0.54 & \textbf{0.85} & 0.39 & 0.84 & 0.83 & 0.83 \\
        
        \bottomrule
    \end{tabular}
    }
    \vspace{-0.5em}
    \caption{Localization performance evaluation in Stanford 2D-3D-S~\cite{stanford2d3d}, compared against PICCOLO (PC)~\cite{piccolo}, structure-based approach (SB), Chamfer distance-based approach (CD), Line Transformer (LT)~\cite{line_transformer}, and CPO~\cite{cpo}.}
    \label{table:stanford}
\vspace{-1em}
\end{table*}

\begin{table}[t]
    \centering
    \resizebox{.7\linewidth}{!}{
    \begin{tabular}{l|cccccc}
        \toprule
        {} & \multicolumn{6}{c}{Accuracy}\\
        Dataset & PC & SB & CD & LT & CPO & LDL \\
        \midrule
        Original & 0.45 & 0.69 & 0.21 & 0.68 & 0.72 & \textbf{0.89} \\
        Gamma & 0.00 & 0.63 & 0.47 & 0.59 & 0.00 & \textbf{0.82} \\
        Intensity & 0.00 & 0.56 & 0.40 & 0.58 & \textbf{0.80} & 0.76 \\
        White Balance & 0.00 & 0.62 & 0.32 & 0.67 & 0.74 & \textbf{0.91} \\
        \bottomrule
    \end{tabular}
    }
    \vspace{-0.5em}
    \caption{Localization accuracy on synthetic color variations applied to Room 3 in the Extreme split from OmniScenes~\cite{piccolo}.}
    \label{table:omniscenes_illum}
\vspace{-0.5em}
\end{table}

\section{Experiments}
\label{sec:exp}
We evaluate LDL in various localization scenarios and analyze its performance.
Our method is mainly implemented using PyTorch~\cite{pytorch}, and is accelerated with a single RTX 2080 GPU.
In all our experiments we set the number of principal directions as $k_{2D}{=}20, k_{3D}{=}3$, the inlier threshold $\tau{=}0.1$, and the number of query points as $|Q|{=}42$.
We report the full hyperparameter setup in the supplementary material.
Following prior works~\cite{piccolo, gosma, gopac}, we report the median translation and rotation errors along with the localization accuracy where a prediction is considered correct if  the translation error is below 0.1m and the rotation error is below 5°.

\paragraph{Datasets}
We evaluate LDL in two indoor localization datasets: Stanford 2D-3D-S~\cite{stanford2d3d} and OmniScenes~\cite{piccolo}.
Stanford-2D-3D-S~\cite{stanford2d3d} contains 1413 panorama images from 272 rooms subdivided into six areas.
Each area has diverse indoor scenes such as offices, hallways, and auditoriums where repetitive structure and featureless regions are present.
OmniScenes contains 4121 panorama images from seven 3D scans, where the panorama images are captured with cameras either handheld or robot mounted, and at different times of day including large changes in furniture configurations.
The dataset has three splits (Robot, Handheld, Extreme) that are recorded in scenes with/without changes, where images in the Extreme split are captured under large camera motion.

\begin{table*}[t]
    \centering
    \resizebox{0.85\linewidth}{!}{
    \begin{tabular}{lc|cccccc|cccccc|cccccc}
        \toprule
        & & \multicolumn{6}{c|}{$t$-error (m)} & \multicolumn{6}{c|}{$R$-error ($^\circ$)} & \multicolumn{6}{c}{Accuracy}\\
        Split & \multicolumn{1}{l|}{Change} & PC & SB & CD & LT & CPO & LDL & PC & SB & CD & LT & CPO & LDL & PC & SB & CD & LT & CPO & LDL \\
        \midrule
        Robot & \xmark & 0.02 & 0.03 & 1.74 & 0.03 & \textbf{0.01} & 0.02 & 0.27 & 0.58 & 89.23 & 0.59 & \textbf{0.12} & 0.49 & 0.69 & \textbf{0.99} & 0.31 & \textbf{0.99} & 0.89 & 0.98 \\
        Hand & \xmark & \textbf{0.01} & 0.03 & 2.10 & 0.03 & \textbf{0.01} & 0.03 & 0.23 & 0.63 & 89.02 & 0.64 & \textbf{0.13} & 0.54 & 0.81 & 0.95 & 0.29 & 0.95 & 0.80 & \textbf{0.97} \\ \midrule
        Robot & \cmark & 1.07 & 0.04 & 1.78 & 0.04 & \textbf{0.02} & 0.03 & 21.03 & 0.64 & 89.27 & 0.65 & 1.46 & \textbf{0.58} & 0.41 & 0.93 & 0.30 & 0.94 & 0.59 & \textbf{0.95} \\
        Hand & \cmark & 0.53 & 0.04 & 1.70 & 0.04 & \textbf{0.02} & 0.03 & 7.54 & 0.71 & 88.50 & 0.70 & \textbf{0.37} & 0.64 & 0.47 & \textbf{0.92} & 0.30 & 0.90 & 0.60 & \textbf{0.92} \\
        Extreme & \cmark & 1.24 & 0.04 & 1.55 & 0.04 & \textbf{0.03} & \textbf{0.03} & 23.71 & 0.83 & 88.54 & 0.84 & \textbf{0.37} & 0.72 & 0.41 & 0.89 & 0.29 & 0.88 & 0.59 & \textbf{0.92} \\
        \bottomrule
    \end{tabular}
    }
    \caption{Localization performance evaluation in OmniScenes~\cite{piccolo}, considering both scenes with and without object layout changes.}
    \label{table:omniscenes}
\end{table*}

\paragraph{Baselines}
We compare LDL against three point-based baselines (PICCOLO, CPO, structure-based) and two line-based baselines (Chamfer distance-based, Line Transformer~\cite{line_transformer}).
PICCOLO (PC)~\cite{piccolo} and the follow-up work CPO~\cite{cpo} is an optimization-based algorithm that finds pose by minimizing the color discrepancy between the point cloud and the query image.
Structure-based approach~\cite{sarlin2019coarse, inloc} (SB) is one of the most prominent methods for visual localization using perspective cameras.
We implement a method for panorama images, where candidate poses are retrieved from an image database using a global feature extractor~\cite{openibl} and further refined using SuperGlue~\cite{sarlin2020superglue} matches.
For fair comparison, we undistort the panorama image into cubemaps and perform feature matching, where the results are then fed to PnP-RANSAC for refinement.
In addition, we construct the database of pose-annotated images by rendering synthetic views at various locations in the colored point cloud.

Chamfer distance-based approach (CD), inspired from Micusik et al.~\cite{line_chamfer}, ranks candidate poses by comparing the spherical Chamfer distance of line segments in the synthetic views against the query image.
Line Transformer by Yoon et al.~\cite{line_transformer} (LT) ranks candidate poses using Transformer-based~\cite{transformer} matching learned for each line segment.
As this baseline also requires a pose-annotated database, we construct a synthetic database similar to the structure-based approach, and apply the undistortion process for fair comparison.
We provide additional details about the baselines in the supplementary material.

\subsection{Localization Evaluation}
\label{sec:loc_eval}
\paragraph{Stanford 2D-3D-S}
We first assess the localization performance of LDL against the baselines in the Stanford 2D-3D-S dataset, as shown in Table~\ref{table:stanford}.
LDL performs competitively against the strong baselines (Structure-based and Line Transformer) that apply powerful neural networks for candidate pose search.
While the dataset contains hallways and auditoriums with large featureless regions or repetitive structure, LDL leverages the holistic distribution of lines using distance functions and shows stable performance without resorting to costly neural network computations.
Further, LDL shows superior performance when compared against the Chamfer distance-based method, which indicates that solely focusing on line matches for ranking candidate poses can lead to suboptimal performance.

\paragraph{OmniScenes}
We additionally compare LDL against baselines in the OmniScenes dataset, as shown in Table~\ref{table:omniscenes}.
Unlike the Stanford 2D-3D-S dataset, all images exhibit blur from camera motion and approximately half of the images contain changes in object layout.
In splits not containing changes, LDL performs competitively against the baselines, which supports our claim that line distance functions enable effective pose search without using neural networks.
Further, LDL attains the highest accuracy in splits containing scene changes and notably in the extreme split that contains the largest amount of motion blur.
This is due to the stable line extraction~\cite{LSD, lineseg_1, lineseg_2, lineseg_3} that enables resilience against motion blur, and the robust distance function comparison (Equation~\ref{eq:loss}) that rejects outliers for handling scene changes.
We further verify the importance of each components in LDL in Section~\ref{sec:perf}.
\begin{figure}[t]
  \centering
    \includegraphics[width=0.95\linewidth]{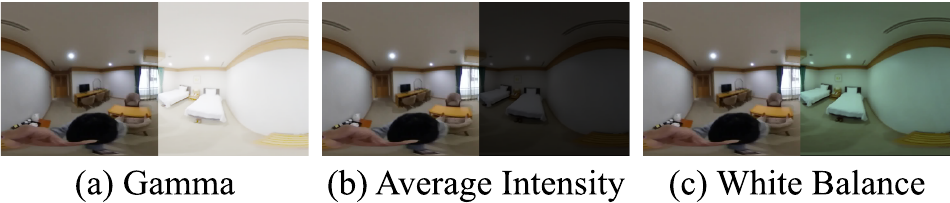}
    \vspace{-0.5em}
    \caption{Color variations for evaluating illumination robustness.}
   \label{fig:color}
\vspace{-1.5em}
\end{figure}

\paragraph{Illumination Robustness Evaluation}
To validate the illumination robustness of LDL, we measure localization performance after applying synthetic color variations.
We select Room 3 from the Extreme split in OmniScenes for evaluation.
As shown in Figure~\ref{fig:color}, the image gamma, white balance, and average intensity are modified to an arbitrary value, where further details  are deferred to the supplementary material.
We report the results of LDL along with the baselines in Table~\ref{table:omniscenes_illum}.
CPO, PICCOLO, and the structure-based baseline all suffer from performance degradation, as the color values are directly utilized for finding initial poses.
Notably, Yoon et al.~\cite{line_transformer} also shows performance drop, as Transformer line features are affected by the illumination changes of the image.
As LDL relies on the spatial structure of line segments for candidate pose search, it is robust to illumination variations, leading to stable performance across all color variations.
Further, note that while all the methods excluding PICCOLO~\cite{piccolo} and CPO~\cite{cpo} use local feature matching for pose refinement, there is a large performance gap between LDL and the other methods.
This validates our focus on designing a stable candidate pose selection method, as modern feature descriptors and matching algorithms~\cite{sarlin2019coarse,sarlin2020superglue,superpoint,d2net} are fairly robust against adversaries such as illumination changes.

\begin{figure}[t]
  \centering
    \includegraphics[width=\linewidth]{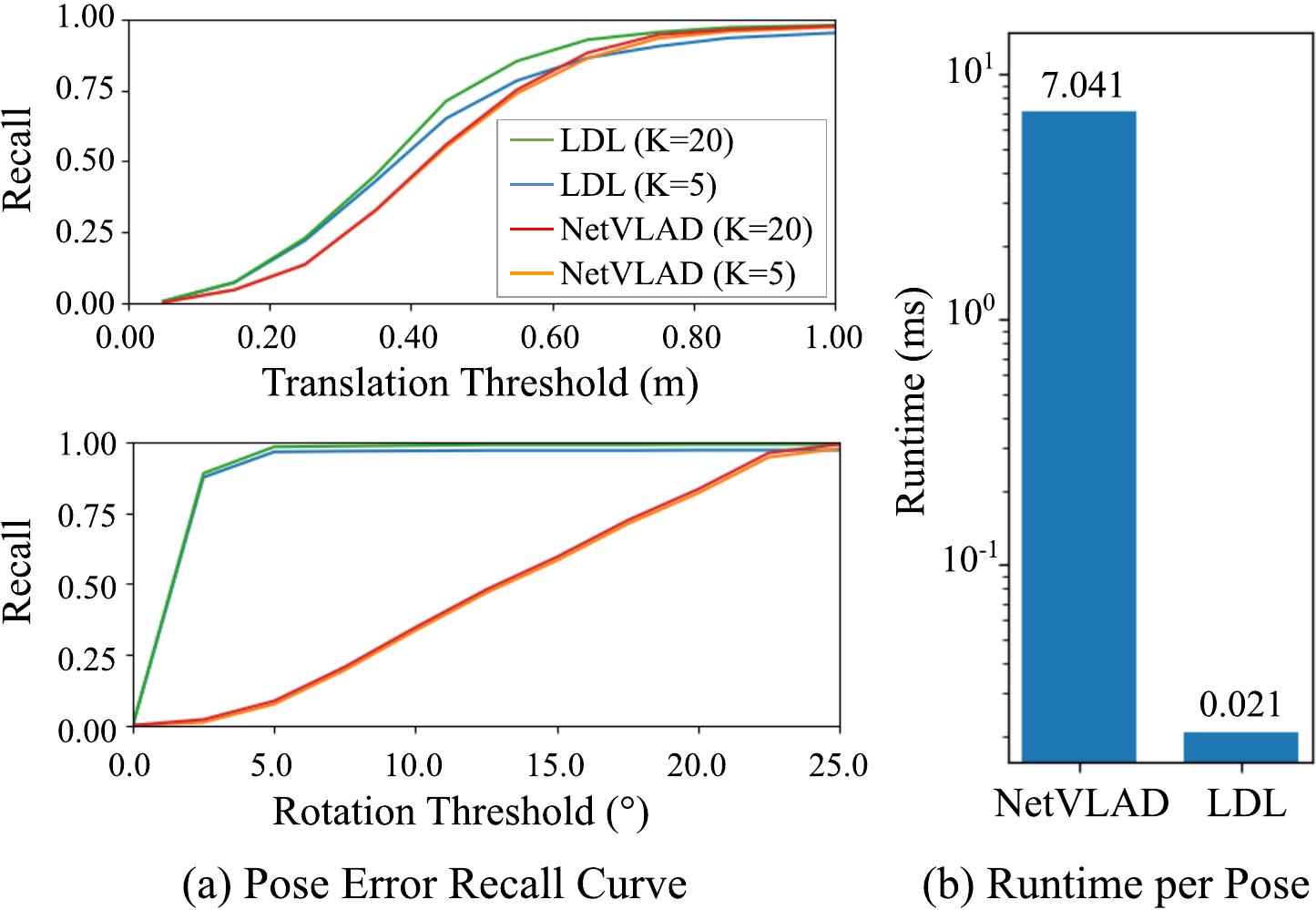}
    \caption{Pose error recall and runtime comparison between candidate pose search using LDL and NetVLAD~\cite{netvlad}.}
   \label{fig:pose_search}
\end{figure}
\begin{table}[t]
    \begin{subtable}{0.45\linewidth}
    \centering
    \resizebox{\linewidth}{!}{
    \begin{tabular}{l|cccc}
        \toprule
        \multirow{2}{*}{Method} & $t$-error & $R$-error & \multirow{2}{*}{Acc.}\\
        &  (m)& ($^\circ$)& \\
        
        \midrule
        SB ($K{=}10$)& 0.06 & 1.18 & 0.63 \\
        SB ($K{=}20$)& \textbf{0.05} & \textbf{1.07} & \textbf{0.71} \\

        \midrule
        LDL ($K{=}10$)& 0.07 & 1.36 & 0.63 \\
        LDL ($K{=}20$)& 0.07 & 1.27 & 0.69 \\
        \bottomrule
    \end{tabular}
    }
    \caption{Multi-Room Localization}
    \end{subtable}
    \Hquad
    \begin{subtable}{0.53\linewidth}
    \centering
    \resizebox{\linewidth}{!}{
    \begin{tabular}{l|cc}
        \toprule
        Component & CPU & GPU \\
        \midrule
        Line Segment Extraction & 0.141 & 0.141 \\
        Rotation Estimation & 1.124 & 0.009 \\
        Distance Function Computation & 0.052 & 0.001 \\
        Candidate Pose Refinement & 5.573 & 0.587 \\
        \midrule
        Total Runtime (sec) & 6.890 & 0.738 \\
        \bottomrule
    \end{tabular}
    }
    \caption{Runtime on CPU and GPU}
    \end{subtable}

    \caption{Multi-room localization compared against Structure-Based method (SB) with various number of candidate poses ($K$) and runtime analysis of LDL.}
    \label{table:runtime_scale}
\vspace{-1.5em}
\end{table}

\subsection{Performance Analysis}
\label{sec:perf}
\paragraph{Candidate Pose Search Evaluation}
To evaluate the efficacy of line distance functions for candidate pose search, we compare the retrieval accuracy of LDL against NetVLAD~\cite{netvlad}, which is a widely used global feature extractor~\cite{sarlin2019coarse,line_transformer,robust_retrieval}.
Note that NetVLAD is used as the candidate pose selection module in the structure-based baseline.
We use the Extreme split from OmniScenes for evaluation, where the translation and rotation error recall curve along with the runtime for processing a single candidate pose is reported in Figure~\ref{fig:pose_search}.
For fair comparison we use the identical pool of translations for both methods as $N_t=50$ and assign a large number of candidate rotations for NetVLAD with $N_r=216$.
Additional setup details are reported in the supplementary material.
While neural network-based pose search methods can perform city-scale search~\cite{netvlad,patchnetvlad,openibl}, the line distance functions in LDL exhibit competitive performance to NetVLAD in indoor environments.
The distance functions provide highly discriminative spatial context, which enables effective pose search.
Furthermore, the runtime for pose search in LDL is much shorter than NetVLAD, due to the highly efficient computation of distance functions only conducted on sparse sphere points. 
This is in contrast to NetVLAD where visual features are computed with a neural network for each view.
The line distance functions enable quick and effective pose initialization, which in turn allow LDL to be usable in various practical localization scenarios.

\paragraph{Runtime Analysis}
We analyze the runtime of LDL in Table~\ref{table:runtime_scale}\textcolor{red}{b} where we decompose the runtime for localizing a single query image from OmniScenes~\cite{piccolo}.
We assume that 3D scanning along with map building is done offline and only consider the computation time for online operations, namely 2D line segment extraction, candidate pose selection and refinement.
Overall, the pose selection process including rotation estimation and distance function computation exhibits a small runtime for both CPU and GPU, which validates the efficiency of our proposed line-based pose search.
Nevertheless, the pose refinement exhibits a relatively larger runtime, which is mainly due to the large number of features in panoramas compared to normal images with a smaller field of view.
While we attained our focus in pose search and used the off-the-shelf local feature matching algorithms for pose refinement~\cite{sarlin2020superglue,superpoint}, devising highly efficient feature matching algorithms tailored specifically for panoramas is left as future work.

\paragraph{Scalability Analysis}
We assess the scalability of LDL to large-scale indoor scenes using the OmniScenes~\cite{piccolo} dataset.
While the previous set of experiments assume room-scale localization scenarios, here we test LDL using the entire OmniScenes dataset as the 3D map.
Table~\ref{table:runtime_scale}\textcolor{red}{a} shows the localization results, where LDL is compared against the structure-based method at various number of candidate poses ($K$).
LDL exhibits performance on a par with the structure-based method, which indicates that line distance functions can scalably handle large scenes consisting of multiple rooms.
Nevertheless, scaling LDL to even larger scale scenes (e.g. building-scale scenes as in InLoc~\cite{inloc}) is left as future work.

\begin{figure}
    \centering
    \includegraphics[width=\linewidth]{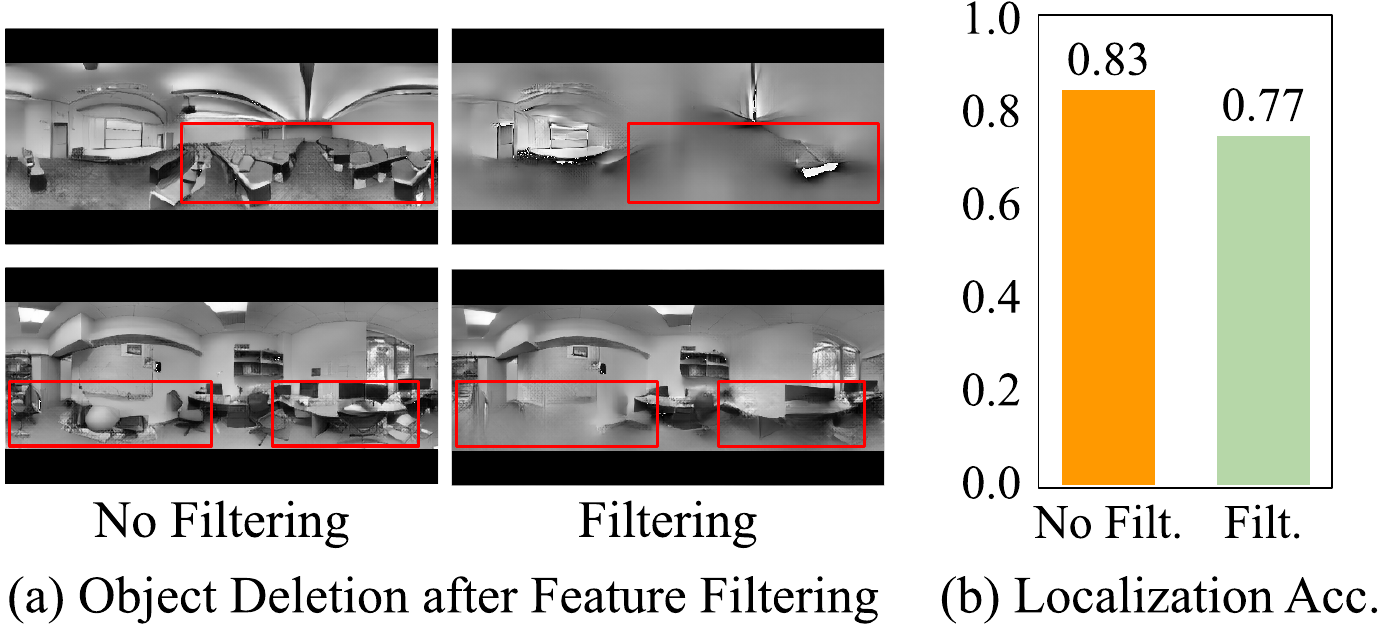}
    \caption{Visualization of feature inversion attacks on panoramic inputs along with the localization accuracy before and after line-based feature filtering}
    \label{fig:privacy}
\end{figure}

\paragraph{Privacy Preservation Analysis}
While the main goal of LDL is to offer fast and robust localization based on lines, we find that with a small modification our method can offer light-weight privacy protection in client-server localization scenarios~\cite{privacy_affine,privacy_analysis,privacy_concern,privacy_line2d,privacy_line3d}.
Following prior works~\cite{ninjadesc,privacy_line2d}, we consider the case where a client using an edge device wants to localize oneself against a 3D line map stored in the cloud.
Privacy breaches occur if the service provider maliciously tries to view the visual data captured by the client.
This is possible even when only the local feature descriptors are shared between the client and server, by using feature inversion methods~\cite{pittaluga_revealing} that reconstruct the original image from a sparse set of local features as shown in Figure~\ref{fig:privacy}.

By changing LDL to only exploit local features near lines during refinement, we can prevent privacy breaches including feature inversion attacks without largely sacrificing localization performance.
First, as LDL uses line segments for candidate pose selection the clients only need to share the extracted line segments with the service providers for initial pose search, instead of the entire view that would be needed for global feature-based methods.
Second, as local features near line segments are shared with the service provider for pose refinement, feature inversion methods cannot faithfully recover the original visual content.
We validate this claim with a small set of experiments performed in the Stanford 2D-3D-S dataset~\cite{stanford2d3d}, where we filter descriptors whose spherical distances to the nearest line segment are over 0.05 rad.
As shown in Figure~\ref{fig:privacy}, this line-based filtering degrades the quality of feature inversion attacks by hiding objects that potentially contain sensitive information while only incurring small drops in localization accuracy.
We report additional details and results regarding the potential of LDL for privacy preservation in the supplementary material.

\subsection{Ablation Study}
We ablate the distance function decomposition, number of query points, and robust loss function, which are key components of LDL in the OmniScenes Extreme split.
In Table~\ref{table:ablation}\textcolor{red}{a}, LDL is first compared against the baseline that does not apply decomposition and use the loss function in Equation~\ref{eq:loss}.
Decomposition leads to a large performance gain, as the distance functions are further disambiguated and split into each principal direction.
We further test the effect of the number of query points $|Q|$ on evaluating the robust loss function.
While increasing the number of query points enhances performance, the improvement is not as significant and incurs additional computation.
Conversely, using a smaller number of query points lead to ambiguities in distance function matching, exhibiting poor performance.
The number of query points $|Q|=42$ balances both the computational efficiency and localization accuracy of LDL.
We finally validate the robust loss function in Equation~\ref{eq:decomp} by comparing LDL against variants using other loss functions: L1, L2, Huber, and Median loss.
Here we report results from the Wedding Hall scene, as this scene contains drastic scene changes with large amounts of outliers.
As shown in Table~\ref{table:ablation}\textcolor{red}{b}, inlier counting proposed in Equation~\ref{eq:decomp} attenuates outliers and exhibits optimal performance, demonstrating the effectiveness of the robust loss function.

\begin{table}[t]
    \begin{subtable}{0.5\linewidth}
    \centering
    \resizebox{\linewidth}{!}{
    \begin{tabular}{l|cccc}
        \toprule
        \multirow{2}{*}{Method} & $t$-error & $R$-error & \multirow{2}{*}{Acc.}\\
        & (m) &($^\circ$) & \\
        \midrule
        w/o Decomposition & 1.00 & 3.97 & 0.37\\
        \midrule
        w/ $|Q|=10$ & 0.04 & 0.85 & 0.77\\
        w/ $|Q|=21$ & 0.04 & 0.71 & 0.88\\
        w/ $|Q|=84$ & \textbf{0.03} & \textbf{0.66} & \textbf{0.95} \\
        \midrule
        Ours ($|Q|=42$)& \textbf{0.03} & 0.72 & 0.92 \\
        \bottomrule
    \end{tabular}
    }
    \caption{Decomposition \& Query Points}
    \end{subtable}
    \Hquad
    \begin{subtable}{0.48\linewidth}
    \centering
    \resizebox{\linewidth}{!}{
    \begin{tabular}{l|ccc}
        \toprule
        \multirow{2}{*}{Method} & $t$-error & $R$-error & \multirow{2}{*}{Acc.}\\
        & (m) &($^\circ$) & \\
        \midrule
        w/ L1 Loss & 0.08 & 1.38 & 0.55 \\
        w/ L2 Loss & 0.17 & 1.48 & 0.34 \\
        w/ Huber Loss & 0.11 & 1.39 & 0.50 \\
        w/ Median Loss & 0.08 & \textbf{1.22} & 0.55 \\
        \midrule
        Ours & \textbf{0.07} & \textbf{1.22} & \textbf{0.68} \\
        \bottomrule
    \end{tabular}
    }
    \caption{Choice of Loss Function}
    \end{subtable}

    \caption{Ablation study of various components of LDL.}
    \label{table:ablation}

\end{table}

\vspace{-0.5em}
\section{Conclusion}
We presented LDL, a fast and robust algorithm for panorama to point cloud localization using line segments.
LDL benefits from the illumination-robustness of line segments and the holistic context of panoramas by using a novel formulation based on line distance functions.
The distance functions effectively handle visual ambiguities of line segments, as they provide spatial meaning to void regions often neglected by existing line-based localization methods.
In addition, by evaluating distance functions only on sparsely sampled query points, LDL performs rapid candidate pose search with accuracy on a par with learning-based global feature extractors.
As a result, LDL performs robust localization in various challenging scenarios with a short runtime.
We expect LDL to complement and enhance the currently prevalent point-based localization algorithms for highly robust and practical localization.

\paragraph{Acknowledgements}
This work was partially supported by the National Research Foundation of Korea(NRF) grant funded by the Korea government(MSIT) (No. RS-2023-00218601), Institute of Information \& communications Technology Planning \& Evaluation (IITP) grant funded by the Korea government(MSIT) (No.2021-0-02068, Artificial Intelligence Innovation Hub), and Samsung Electronics Co., Ltd.
Young Min Kim is the corresponding author.

\appendix
\renewcommand\thetable{\thesection.\arabic{table}}    
\setcounter{table}{0}
\renewcommand\thefigure{\thesection.\arabic{figure}}    
\setcounter{figure}{0}

\pagenumbering{gobble}

\section{Details on LDL}

\paragraph{Principal Direction Computation}
We explain the details of principal direction computation.
Recall that the principal directions in 2D and 3D are defined as the top $k_{2D}$ and $k_{3D}$ most common line directions.
The 2D principal directions are extracted from vanishing points.
When parallel lines are projected on an image, they appear to converge at a point, which is referred to as a vanishing point.
To locate vanishing points, we extrapolate detected line segments and find their intersections.
Since we are using panoramic images, we use spherical projection of lines and vanishing points.
Specifically, we create a uniform spherical grid and count the number of intersection points in each grid cell, which we referred to as `voting' in the main paper.
We select the top $k_{2D}$ grid locations with the most votes as the 2D principal directions.
For 3D principal directions, we similarly aggregate votes for 3D line directions and extract the top $k_{3D}$ votes.
Note that we fix the filtering parameters for all our experiments and LDL achieves competitive results.

\begin{figure}[t]
  \centering
    \includegraphics[width=0.9\linewidth]{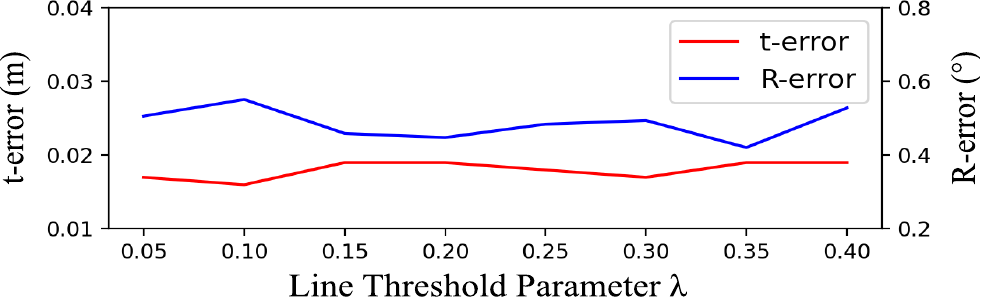}
   \caption{Localization error against line threshold parameter $\lambda$.}
   \label{fig:reb_thres}
\vspace{-1em}
\end{figure}

\paragraph{Line Filtering}
Prior to localization, recall from Section 3.1 that LDL filters short lines.
Specifically, given the point cloud with the bounding box size of $b_x \times b_y \times b_z$, we filter out 3D line segments shorter than $\lambda (b_x+b_y+b_z)/3$, where $\lambda=0.1$ in all our experiments.
The 2D line segments are then filtered to match the filtering rate of 3D line segments.
Note the threshold parameter $\lambda$ does not play a critical role in performance.
Figure~\ref{fig:reb_thres} shows the median localization error measured in Room 1 from OmniScenes~[\textcolor{green}{5}].
The errors are nearly constant with respect to varying $\lambda$.

\begin{figure}[t]
  \centering
  \includegraphics[width=.7\linewidth]{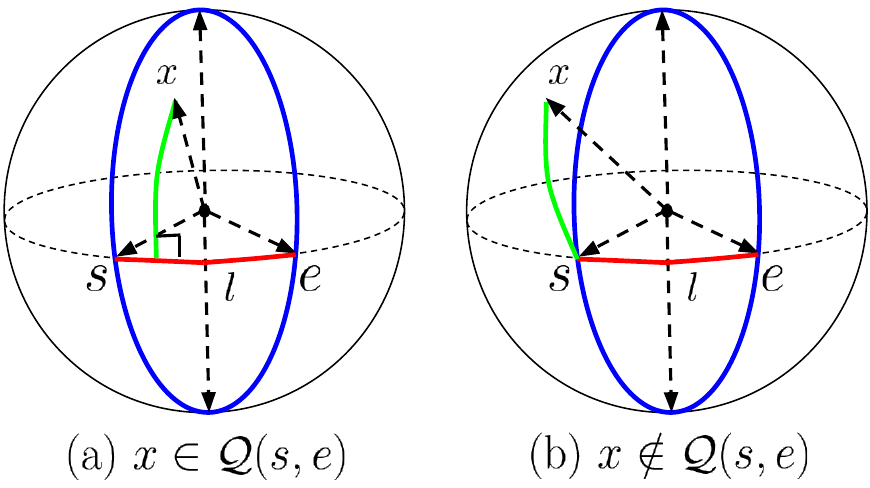}
   \caption{Given a line segment $l$ (red), the distance (green) from point $x$ to the $l$ is defined depending on whether $x$ lies on the spherical quadrilateral (blue) $\mathcal{Q}(s, e)$.}
   \label{fig:squad}
\vspace{-0.5em}
\end{figure}

\paragraph{Spherical Quadrilateral for Computing Line Distance Functions}
We illustrate the spherical quadrilateral used for computing distance functions from Section 3.
As shown in Figure~\ref{fig:squad}, given a line segment $l$ on a sphere with a start point $s$ and an end point $e$, the spherical quadrilateral $\mathcal{Q}(s,e)$ is formed by connecting $\{s, e, \pm (s\times e) / \|s\times e\|\}$.
The spherical quadrilateral is used in Equation 2 to compute the distance $D(x, l)$ from a point $x$ to a line segment $l$ on a sphere.
Here, $D(x, l)$ is computed differently depending on whether $x$ lies on $\mathcal{Q}(s,e)$.
The 2D and 3D line distance functions (Equation 1, 3) are further built upon this definition of $D(x,l)$.

\paragraph{Hyperparameter Setup}
Here we report the hyperparameter setup of LDL.
As explained in Section 3, from $N_t \times N_r$ poses we select $K$ candidate poses by comparing the distance functions with the robust lost function in Equation 5.
Recall that we use the candidate rotation estimation step in Section 3.2 to choose $N_r$ rotations.
For the $N_t$ translations, we follow the design choice of prior works~\cite{piccolo, cpo, gosma, gopac} and employ uniform grid partitions for Stanford2D-3D-S~\cite{stanford2d3d} and centroids of octrees as in Rodenberg \etal~\cite{octree} for OmniScenes~\cite{piccolo}.
We set $K{=}20, N_t{=}800$ for OmniScenes~\cite{piccolo} and $K{=}20, N_t{=}1700$ for Stanford 2D-3D-S~\cite{stanford2d3d}.
We use an increased number of translations for Stanford 2D-3D-S to cope with large scenes such as auditoriums and hallways.
Nevertheless, note that LDL can quickly search promising candidate poses: even in Stanford 2D-3D-S candidate pose search finishes within 0.02 seconds.

\begin{figure}[t]
  \centering
    \includegraphics[width=\linewidth]{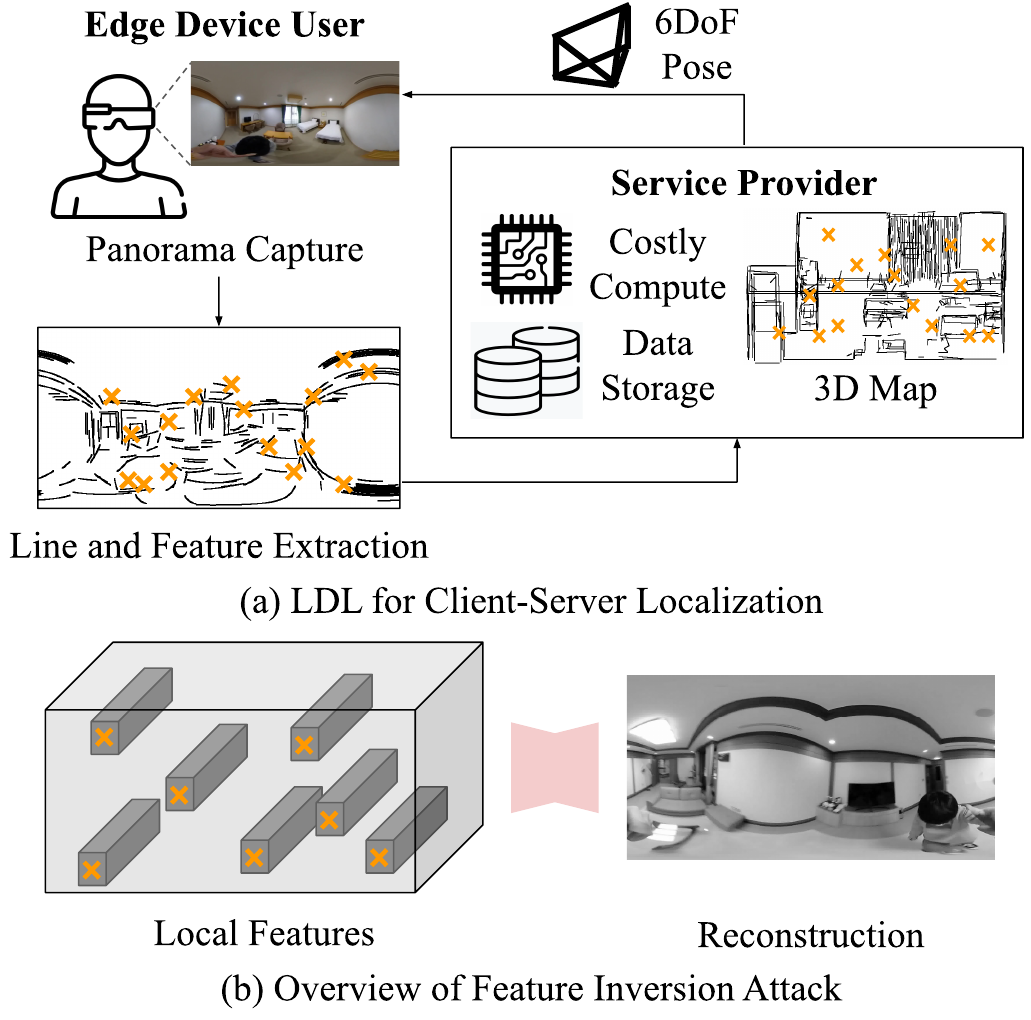}
   \caption{Client-server localization setup using LDL. (a) The edge device user captures the raw 2D data and shares the lines and local features near lines with the service provider. The service provider provides the 6DoF pose using the shared information along with the 3D map. (b) While the service provider can attempt feature inversion attacks by training neural networks that learn image reconstructions from local feature inputs, this cannot fully recover the sensitive visual details for LDL as only a fraction of information is shared. }
   \label{fig:server}
\vspace{-1em}
\end{figure}

\paragraph{Potential for Privacy Preservation}
As explained in Section 4.2, while the primary goal of LDL is to offer fast and robust localization, our approach can also be extended to offer low cost protection against various privacy breaches in client-server localization.
To cope with edge devices having limited computing power, modern location-based services employ a client-server localization setup~\cite{privacy_analysis,privacy_concern} where the visual data of the edge device is shared with the service provider~\cite{privacy_analysis, pittaluga_revealing}.
Based on the shared information, the service provider performs the actual localization pipeline and returns the estimated 6DoF pose to the edge device user.

We adapt LDL to the client-server localization scenario while offering privacy protection by having the edge device user to only share lines and local features near lines during localization.
Specifically, as shown in Figure~\ref{fig:server}, we modify the pose refinement phase of LDL to operate using local features near lines, instead of all the visible local features used for the original refinement explained in Section 3.4.
Here we only consider line segments whose lengths are over a designated threshold as explained in Section 3.1 and directions are parallel to one of the 2D principal directions.
Such a modification results in privacy protection against feature inversion attacks ~\cite{privacy_analysis,pittaluga_revealing,ninjadesc}, which take local feature vectors as input and outputs an image reconstruction.
Note that LDL naturally offers privacy protection during pose selection as it only uses line segments for this phase and thus does not necessitate the clients to share their entire view with the service provider.
We further demonstrate the potential of LDL for privacy protection through experiments shown in Section~\ref{sec:additional_privacy}.

\section{Additional Experimental Results}
\label{sec:ablation}
\setcounter{table}{0}
\begin{table}[t]
    \centering
    \resizebox{.75\linewidth}{!}{
    \setlength{\tabcolsep}{3pt}
    \begin{tabular}{l|cc|cc|cc}
        \toprule
        {} & \multicolumn{2}{c|}{$t$-error (m)} & \multicolumn{2}{c|}{$R$-error ($^\circ$)} & \multicolumn{2}{c}{Accuracy}\\
        Area & LDL & LDL\textsuperscript{LS} & LDL & LDL\textsuperscript{LS} & LDL & LDL\textsuperscript{LS} \\
        \midrule
        Area 1 & \textbf{0.02} & \textbf{0.02} & \textbf{0.54} & 0.60 & \textbf{0.86} & 0.75 \\
        Area 2 & \textbf{0.02} & 0.05 & \textbf{0.66} & 0.79 & \textbf{0.77} & 0.57 \\
        Area 3 & \textbf{0.02} & 0.03 & \textbf{0.54} & 0.73 & \textbf{0.89} & 0.69 \\
        Area 4 & \textbf{0.02} & \textbf{0.02} & \textbf{0.48} & 0.57 & \textbf{0.88} & 0.72 \\
        Area 5 & \textbf{0.02} & 0.03 & \textbf{0.54} & 0.61 & \textbf{0.81} & 0.59 \\
        Area 6 & \textbf{0.02} & \textbf{0.02} & \textbf{0.50} & 0.58 & \textbf{0.83} & 0.66 \\
        \midrule
        Total & \textbf{0.02} & 0.03 & \textbf{0.53} & 0.64 & \textbf{0.83} & 0.66 \\

        \bottomrule
    \end{tabular}
    }
    \caption{Ablation study of uniformly sampling query points on the unit sphere. LDL is compared against a variant using query points sampled along 2D line segment locations (LDL\textsuperscript{LS}) in the Stanford 2D-3D-S dataset~\cite{stanford2d3d}.}

    \label{table:stanford}
\end{table}

\subsection{Additional Ablation Study}
\paragraph{Choice of Query Point Locations} 
We report the impact of choosing uniformly sampled query points for evaluating distance functions.
Recall that we rank $N_t \times N_r$ poses with the robust loss function in Equation 5, where the query points $Q$ are uniformly sampled from a unit sphere.
We compare LDL against a variant that uses query points sampled along the 2D line segment locations.
Namely, this variant only considers regions with line segments, in contrast to LDL that equally considers regions lacking lines.

We make quantitative evaluations between LDL and the variant using the Stanford 2D-3D-S~\cite{stanford2d3d} dataset.
For fair comparison, we use identical hyperparameters as the original implementation of LDL.
As shown in Table~\ref{table:stanford}, uniform sampling employed in LDL leads to large amounts of performance improvement.
By fairly using all regions on the sphere, LDL effectively utilizes the spatial context from the line distance functions and performs effective localization.

\begin{table}[t]
    \centering
    \resizebox{0.7\linewidth}{!}{
    \begin{tabular}{l|ccc}
        \toprule
        Method & $t$-error (m) & $R$-error ($^\circ$) & Acc.\\
        \midrule
        w/ L1 Loss & 0.08 & 1.38 & 0.55 \\
        w/ L2 Loss & 0.17 & 1.48 & 0.34 \\
        w/ Huber Loss & 0.11 & 1.39 & 0.50 \\
        w/ Median Loss & 0.08 & \textbf{1.22} & 0.55 \\
        \midrule
        Ours & \textbf{0.07} & \textbf{1.22} & \textbf{0.68} \\
        \bottomrule
    \end{tabular}
    }
    \caption{Ablation study on the choice of loss functions evaluated in OmniScenes~\cite{piccolo}.}
    \label{table:loss_ablation}
\end{table}

\paragraph{Choice of Loss Function}
We validate the robust loss function in Equation 5 by comparing LDL against variants using other loss functions: L1, L2, Huber, and Median loss.
Here we report results from the Wedding Hall scene in OmniScenes, as this scene contains drastic scene changes with large amounts of outliers.
As shown in Table~\ref{table:loss_ablation}, the inlier counting proposed in Equation 5 attenuates outliers in the Extreme split and exhibits optimal performance, demonstrating the effectiveness of the robust loss function.

\begin{figure}[t]
  \centering
    \includegraphics[width=0.8\linewidth]{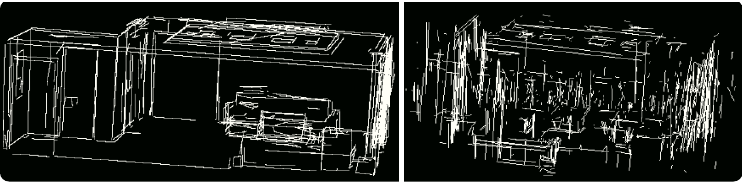}
   \caption{3D Lines from 3D Scanning (Left) and SfM (Right).}
   \label{fig:reb_line}
\end{figure}
\begin{table}[t]
\centering
    \resizebox{0.55\linewidth}{!}{
    \begin{tabular}{l|ccc}
        \toprule
        Method & $t$-error (m) & $R$-error ($^\circ$) & Acc.\\
        \midrule
        SfM & \textbf{0.03} & 0.80 & 0.85\\
        3D Scan & \textbf{0.03} & \textbf{0.71} & \textbf{0.98} \\
        \bottomrule
    \end{tabular}
    }
    \caption{Evaluation results of LDL on noisier line maps obtained using structure from motion and Line3D++~\cite{line3dpp}.}
    \label{table:noisy_line}

\end{table}

\subsection{Evaluation in Noisier Maps}
In the main paper, we extract 3D lines from point clouds obtained using Matterport 3D scanners~\cite{matterport_scan}.
Here we run LDL on noisier line maps created using structure-from-motion (SfM) and Line3D++~\cite{line3dpp}.
As shown in Figure~\ref{fig:reb_line}, the maps are more noisier than those from 3D scans.
Table~\ref{table:noisy_line} shows the localization results from Room 3, 5 in Omniscenes under different types of line maps (note the new pipeline did not produce reliable maps in other scenes).
Even though LDL was run with the exact same hyperparameters as in the main paper, it shows only a small amount of performance drop, which indicates that it can robustly handle noisier SfM-based line maps which are generated without 3D scanners.

\begin{figure}[t]
  \centering
    \includegraphics[width=0.9\linewidth]{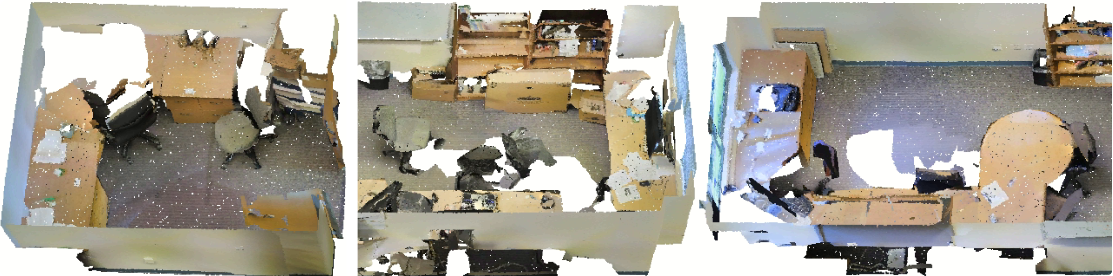}
   \caption{Top-down view of offices in Stanford 2D-3D-S~[\textcolor{green}{3}].}
   \label{fig:reb_office}
\end{figure}
\begin{table}[t]
\centering
    \resizebox{0.6\linewidth}{!}{
    \begin{tabular}{l|ccc}
        \toprule
        Method & $t$-error (m) & $R$-error ($^\circ$) & Acc.\\
        \midrule
        LDL & \textbf{0.02} & \textbf{0.54} & \textbf{0.90}\\
        Structure-Based & 0.03 & 0.58 & 0.89 \\
        \bottomrule
    \end{tabular}
    }
    \caption{Evaluation on Large Scale Scenes }
    \label{table:reb_iccv_scale}
\end{table}

\subsection{Additional Evaluation in Large-Scale Maps}
In the main paper we demonstrated that LDL can perform competitively against the structure-based method in large scenes by testing multiple room localization in OmniScenes~[\textcolor{green}{23}].
To further show the scalability of LDL, we evaluate on 20 office rooms from Stanford 2D-3D-S~[\textcolor{green}{3}], and localize each image against the jointly composed 3D map.
The 20 office rooms contain similar structures, as shown in Figure~\ref{fig:reb_office}.
Even in such conditions, LDL shows similar performance against the structure-based method as shown in Table~\ref{table:reb_iccv_scale}.
While scalability has not been the main goal of this paper, LDL shows the potential to be deployed in large-scale localization settings containing visual ambiguities.

\begin{table}[t]
    \begin{subtable}{\linewidth}
    \centering
    \resizebox{0.9\linewidth}{!}{
\begin{tabular}{l|cccccc}
\toprule
Accuracy (0.05 m, $5^\circ$) &  PC &  CPO &  SB &  LT &  CD &  LDL \\
\midrule
Robot & 0.66 & 0.88 & 0.86 & 0.85 & 0.27 & \textbf{0.92} \\
Hand & 0.77 & 0.77 & 0.73 & 0.72 & 0.22 & \textbf{0.82} \\
Change Robot & 0.39 & 0.58 & 0.72 & 0.72 & 0.21 & \textbf{0.78} \\
Change Hand & 0.45 & 0.58 & 0.68 & 0.70 & 0.22 & \textbf{0.72} \\
Extreme & 0.38 & 0.57 & 0.63 & 0.62 & 0.20 & \textbf{0.71} \\
\bottomrule
\end{tabular}
}
    \caption{Accuracy at translation and rotation threshold 0.05 m, $5^\circ$}
    \end{subtable}

    \smallskip
    \begin{subtable}{\linewidth}
    \centering
    \resizebox{0.9\linewidth}{!}{
\begin{tabular}{l|cccccc}
\toprule
Accuracy (0.05 m, $10^\circ$) & PC &  CPO & SB & LT & CD &  LDL \\
\midrule
Robot & 0.66 & 0.88 & 0.86 & 0.85 & 0.27 & \textbf{0.92} \\
Hand & 0.77 & 0.77 & 0.73 & 0.72 & 0.22 & \textbf{0.82} \\
Change Robot & 0.39 & 0.58 & 0.72 & 0.72 & 0.21 & \textbf{0.78} \\
Change Hand & 0.45 & 0.58 & 0.68 & 0.70 & 0.22 & \textbf{0.72} \\
Extreme & 0.38 & 0.57 & 0.63 & 0.62 & 0.20 & \textbf{0.71} \\
\bottomrule
\end{tabular}    
    }
    \caption{Accuracy at translation and rotation threshold 0.05 m, $10^\circ$}
    \end{subtable}

    \begin{subtable}{\linewidth}
    \centering
    \resizebox{0.9\linewidth}{!}{
\begin{tabular}{l|cccccc}
\toprule
Accuracy (0.1 m, $5^\circ$) &  PC &  CPO &  SB &  LT &  CD &  LDL \\
\midrule
Robot & 0.69 & 0.89 & \textbf{0.99} & \textbf{0.99} & 0.31 & \textbf{0.98} \\
Hand & 0.81 & 0.80 & 0.95 & 0.95 & 0.29 & \textbf{0.97} \\
Change Robot & 0.41 & 0.59 & 0.93 & 0.94 & 0.30 & \textbf{0.95} \\
Change Hand & 0.47 & 0.60 & \textbf{0.92} & 0.90 & 0.30 & \textbf{0.92} \\
Extreme & 0.41 & 0.59 & 0.89 & 0.88 & 0.29 & \textbf{0.92} \\
\bottomrule
\end{tabular}

    }
    \caption{Accuracy at translation and rotation threshold 0.1 m, $5^\circ$}
    \end{subtable}

    \begin{subtable}{\linewidth}
    \centering
    \resizebox{0.9\linewidth}{!}{
\begin{tabular}{l|cccccc}
\toprule
Accuracy (0.1 m, $10^\circ$) & PC &  CPO & SB & LT & CD &  LDL \\
\midrule
Robot & 0.69 & 0.89 & \textbf{0.99} & \textbf{0.99} & 0.32 & \textbf{0.98} \\
Hand & 0.81 & 0.80 & 0.95 & 0.95 & 0.29 & \textbf{0.97} \\
Change Robot & 0.41 & 0.59 & 0.93 & 0.94 & 0.30 & \textbf{0.95} \\
Change Hand & 0.47 & 0.60 & \textbf{0.92} & 0.90 & 0.30 & \textbf{0.92} \\
Extreme & 0.41 & 0.59 & 0.89 & 0.88 & 0.29 & \textbf{0.92} \\
\bottomrule
\end{tabular}    
    }
    \caption{Accuracy at translation and rotation threshold 0.1 m, $10^\circ$}
    \end{subtable}

    \begin{subtable}{\linewidth}
    \centering
    \resizebox{0.9\linewidth}{!}{
\begin{tabular}{l|cccccc}
\toprule
Accuracy (0.2 m, $5^\circ$) & PC &  CPO & SB & LT & CD &  LDL \\
\midrule
Robot & 0.70 & 0.89 & \textbf{1.00} & \textbf{1.00} & 0.34 & \textbf{0.99} \\
Hand & 0.81 & 0.81 & 0.98 & 0.98 & 0.32 & \textbf{0.99} \\
Change Robot & 0.41 & 0.59 & 0.98 & \textbf{0.99} & 0.33 & \textbf{0.98} \\
Change Hand & 0.48 & 0.60 & \textbf{0.97} & \textbf{0.97} & 0.34 & \textbf{0.97} \\
Extreme & 0.42 & 0.60 & 0.96 & 0.96 & 0.34 & \textbf{0.98} \\
\bottomrule
\end{tabular}    
    }
    \caption{Accuracy at translation and rotation threshold 0.2 m, $5^\circ$}
    \end{subtable}

    \begin{subtable}{\linewidth}
    \centering
    \resizebox{0.9\linewidth}{!}{
\begin{tabular}{l|cccccc}
\toprule
Accuracy (0.2 m, $10^\circ$) & PC &  CPO & SB & LT & CD &  LDL \\
\midrule
Robot & 0.70 & 0.89 & \textbf{1.00} & \textbf{1.00} & 0.34 & \textbf{0.99} \\
Hand & 0.81 & 0.81 & 0.98 & 0.98 & 0.33 & \textbf{0.99} \\
Change Robot & 0.41 & 0.59 & 0.98 & \textbf{0.99} & 0.33 & \textbf{0.98} \\
Change Hand & 0.49 & 0.60 & \textbf{0.97} & \textbf{0.97} & 0.34 & \textbf{0.97} \\
Extreme & 0.42 & 0.60 & 0.96 & 0.96 & 0.34 & \textbf{0.98} \\
\bottomrule
\end{tabular}    
    }
    \caption{Accuracy at translation and rotation threshold 0.2 m, $10^\circ$}
    \end{subtable}

\caption{Localization accuracy at various thresholds in the OmniScenes~\cite{piccolo} dataset.}
    \label{table:full_results}
\vspace{-1.5em}
\end{table}
\begin{table}[t]
    \begin{subtable}{\linewidth}
    \centering
    \resizebox{0.9\linewidth}{!}{
\begin{tabular}{l|cccccc}
\toprule
Accuracy (0.05 m, $5^\circ$) &  PC &  CPO &  SB &  LT &  CD &  LDL \\
\midrule
Area 1 & 0.66 & \textbf{0.89} & 0.83 & 0.83 & 0.46 & 0.83 \\
Area 2 & 0.42 & \textbf{0.81} & 0.63 & 0.63 & 0.30 & 0.69 \\
Area 3 & 0.53 & 0.76 & 0.81 & 0.82 & 0.34 & \textbf{0.86} \\
Area 4 & 0.48 & 0.83 & 0.87 & \textbf{0.88} & 0.43 & 0.85 \\
Area 5 & 0.44 & 0.73 & 0.68 & 0.69 & 0.34 & \textbf{0.74} \\
Area 6 & 0.68 & \textbf{0.90} & 0.80 & 0.82 & 0.45 & 0.81 \\
\bottomrule
\end{tabular}
}
    \caption{Accuracy at translation and rotation threshold 0.05 m, $5^\circ$}
    \end{subtable}

    \smallskip
    \begin{subtable}{\linewidth}
    \centering
    \resizebox{0.9\linewidth}{!}{
\begin{tabular}{l|cccccc}
\toprule
Accuracy (0.05 m, $10^\circ$) & PC &  CPO & SB & LT & CD &  LDL \\
\midrule
Area 1 & 0.66 & \textbf{0.90} & 0.83 & 0.83 & 0.46 & 0.83 \\
Area 2 & 0.42 & \textbf{0.81} & 0.63 & 0.63 & 0.30 & 0.69 \\
Area 3 & 0.53 & 0.76 & 0.81 & 0.82 & 0.34 & \textbf{0.86} \\
Area 4 & 0.48 & 0.83 & 0.87 & \textbf{0.88} & 0.43 & 0.85 \\
Area 5 & 0.44 & 0.73 & 0.68 & 0.69 & 0.34 & \textbf{0.74} \\
Area 6 & 0.68 & \textbf{0.90} & 0.80 & 0.82 & 0.45 & 0.81 \\ 
\bottomrule
\end{tabular}    
    }
    \caption{Accuracy at translation and rotation threshold 0.05 m, $10^\circ$}
    \end{subtable}

    \begin{subtable}{\linewidth}
    \centering
    \resizebox{0.9\linewidth}{!}{
\begin{tabular}{l|cccccc}
\toprule
Accuracy (0.1 m, $5^\circ$) &  PC &  CPO &  SB &  LT &  CD &  LDL \\
\midrule
Area 1 & 0.66 & \textbf{0.90} & 0.89 & \textbf{0.90} & 0.50 & 0.86 \\
Area 2 & 0.45 & \textbf{0.81} & 0.76 & 0.74 & 0.35 & 0.77 \\
Area 3 & 0.57 & 0.78 & \textbf{0.92} & 0.88 & 0.36 & \textbf{0.89} \\
Area 4 & 0.49 & 0.83 & \textbf{0.91} & \textbf{0.91} & 0.46 & 0.88 \\
Area 5 & 0.44 & 0.74 & 0.80 & 0.79 & 0.36 & \textbf{0.81} \\
Area 6 & 0.69 & \textbf{0.90} & 0.88 & 0.87 & 0.47 & 0.83 \\
\bottomrule
\end{tabular}

    }
    \caption{Accuracy at translation and rotation threshold 0.1 m, $5^\circ$}
    \end{subtable}

    \begin{subtable}{\linewidth}
    \centering
    \resizebox{0.9\linewidth}{!}{
\begin{tabular}{l|cccccc}
\toprule
Accuracy (0.1 m, $10^\circ$) & PC &  CPO & SB & LT & CD &  LDL \\
\midrule
Area 1 & 0.66 & \textbf{0.90} & 0.89 & \textbf{0.90} & 0.50 & 0.86 \\
Area 2 & 0.45 & \textbf{0.81} & 0.76 & 0.74 & 0.35 & 0.77 \\
Area 3 & 0.57 & 0.78 & \textbf{0.92} & 0.88 & 0.36 & \textbf{0.89} \\
Area 4 & 0.49 & 0.83 & \textbf{0.91} & \textbf{0.91} & 0.46 & 0.88 \\
Area 5 & 0.44 & 0.74 & 0.80 & 0.79 & 0.36 & \textbf{0.81} \\
Area 6 & 0.69 & \textbf{0.90} & 0.88 & 0.87 & 0.47 & 0.83 \\
\bottomrule
\end{tabular}    
    }
    \caption{Accuracy at translation and rotation threshold 0.1 m, $10^\circ$}
    \end{subtable}

    \begin{subtable}{\linewidth}
    \centering
    \resizebox{0.9\linewidth}{!}{
\begin{tabular}{l|cccccc}
\toprule
Accuracy (0.2 m, $5^\circ$) & PC &  CPO & SB & LT & CD &  LDL \\
\midrule
Area 1 & 0.67 & \textbf{0.90} & 0.89 & \textbf{0.90} & 0.50 & 0.86 \\
Area 2 & 0.47 & \textbf{0.81} & 0.80 & \textbf{0.81} & 0.37 & 0.78 \\
Area 3 & 0.59 & 0.81 & \textbf{0.96} & 0.93 & 0.41 & \textbf{0.95} \\
Area 4 & 0.50 & 0.83 & \textbf{0.94} & \textbf{0.93} & 0.47 & 0.89 \\
Area 5 & 0.47 & 0.78 & \textbf{0.84} & \textbf{0.84} & 0.39 & \textbf{0.84} \\
Area 6 & 0.69 & \textbf{0.90} & 0.88 & 0.88 & 0.48 & 0.84 \\
\bottomrule
\end{tabular}    
    }
    \caption{Accuracy at translation and rotation threshold 0.2 m, $5^\circ$}
    \end{subtable}

    \begin{subtable}{\linewidth}
    \centering
    \resizebox{0.9\linewidth}{!}{
\begin{tabular}{l|cccccc}
\toprule
Accuracy (0.2 m, $10^\circ$) & PC &  CPO & SB & LT & CD &  LDL \\
\midrule
Area 1 & 0.67 & \textbf{0.90} & 0.89 & \textbf{0.90} & 0.50 & 0.86 \\
Area 2 & 0.47 & \textbf{0.81} & 0.80 & \textbf{0.81} & 0.37 & 0.78 \\
Area 3 & 0.59 & 0.81 & \textbf{0.96} & 0.93 & 0.41 & \textbf{0.95} \\
Area 4 & 0.50 & 0.83 & \textbf{0.94} & \textbf{0.93} & 0.47 & 0.89 \\
Area 5 & 0.47 & 0.78 & \textbf{0.84} & \textbf{0.84} & 0.39 & \textbf{0.84} \\
Area 6 & 0.69 & \textbf{0.90} & 0.88 & 0.88 & 0.48 & 0.84 \\
\bottomrule
\end{tabular}    
    }
    \caption{Accuracy at translation and rotation threshold 0.2 m, $10^\circ$}
    \end{subtable}

\caption{Localization accuracy at various thresholds in the Stanford 2D-3D-S~\cite{piccolo} dataset.}
    \label{table:full_results_stanford}
\vspace{-1.5em}
\end{table}

\subsection{Full Localization Evaluation Results at Various Accuracy Thresholds}
We share the full localization results for the OmniScenes~\cite{piccolo} and Stanford 2D-3D-S~\cite{stanford2d3d} datasets in Table~\ref{table:full_results},~\ref{table:full_results_stanford}.
Here we additionally show the localization accuracy at various accuracy thresholds.
Our method can perform competitively against all the tested baselines across various thresholds, while performing light-weight pose search with line distance functions.

\subsection{Privacy Preservation Results}
\label{sec:additional_privacy}
\begin{table}[t]
\begin{subtable}{\linewidth}
    \centering
    \resizebox{0.65\linewidth}{!}{
    \begin{tabular}{l|cccc}
        \toprule
        Method & $t$-error (m) & $R$-error ($^\circ$) & Acc.\\
        \midrule
        No Filtering & \textbf{0.02} & \textbf{0.53} & \textbf{0.83} \\
        Filtering & 0.03 & 0.64 & 0.77 \\
        \bottomrule
    \end{tabular}
    }
\caption{Localization Performance Evaluation in Stanford 2D-3D-S~\cite{stanford2d3d}}
\end{subtable}

\smallskip

\begin{subtable}{\linewidth}
    \centering
    \resizebox{0.65\linewidth}{!}{
    \begin{tabular}{l|cccc}
        \toprule
        Method & 20-PSNR & 1-SSIM & MAE  \\
        \midrule
        No Filtering & 1.1717 & 0.5505 & 0.1598 \\
        Filtering & \textbf{1.7577} & \textbf{0.6027} & \textbf{0.1773} \\
        \bottomrule
    \end{tabular}
    }
\caption{Reconstruction Quality of Feature Inversion Attacks}
\end{subtable}

\smallskip

\caption{Privacy-preservation evaluation of modified LDL using line-based feature filtering evaluated in Stanford 2D-3D-S dataset~\cite{stanford2d3d}. The simple filtering incurs only a small drop in localization accuracy while largely increasing the image error metrics.}
\label{table:full_preservation}
\end{table}
\begin{figure}[t]
  \centering
    \includegraphics[width=\linewidth]{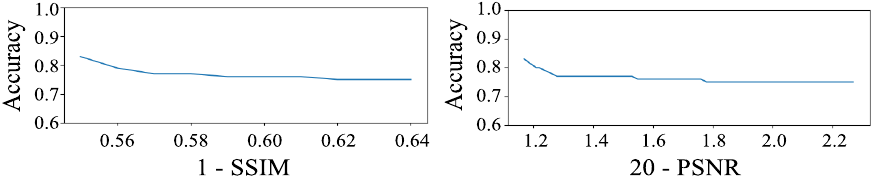}
   \caption{Privacy-utility curve drawn from various values of line-based filtering thresholds in the Stanford 2D-3D-S dataset. While the reconstruction quality of feature inversion attacks largely degrade as we filter out more feature points, the localization accuracy remains relatively constant.}
   \label{fig:barplot_additional}
\vspace{-1em}
\end{figure}
We share the detailed privacy evaluation results on the Stanford 2D-3D-S~\cite{obj_360} dataset.
Table~\ref{table:full_preservation} shows the localization accuracy along with the feature inversion attack results.
The image error metrics (20 - PSNR, 1-SSIM, MAE) of the feature inversion attacks measured against the original panorama consistently increase for all tested scenarios, indicating that our line-based feature filtering can successfully hide visual details.
This notion is further verified through the additional qualitative samples in Figure~\ref{fig:qualitative_additional} where the sensitive visual data such as tabletop clutter are removed after filtering.
Nevertheless, note that the filtering process only incurs a small drop in localization performance.
We finally evaluate how LDL balances privacy (feature inversion protection) and utility (localization accuracy) while using line-based feature filtering.
In Figure~\ref{fig:barplot_additional} we plot the image error metrics of feature inversion attacks against the original image along with the localization accuracy using various line-based filtering threshold values.
While the discrepancy values increase largely, the localization accuracy remains relatively constant.
Thus the modified version of LDL can balance between privacy protection and accurate localization, suggesting its future potential as a robust privacy-preserving localization algorithm.

Nevertheless, the current modification cannot fully hide keypoints from large structures such as walls and ceilings.
While these regions typically do not contain sensitive visual information, some users may want their entire views to be hidden from service providers.
Developing a more secure line-based localization algorithm that could alleviate a wider range of concerns is left as future work.

\begin{figure*}[t]
  \centering
    \includegraphics[width=0.95\linewidth]{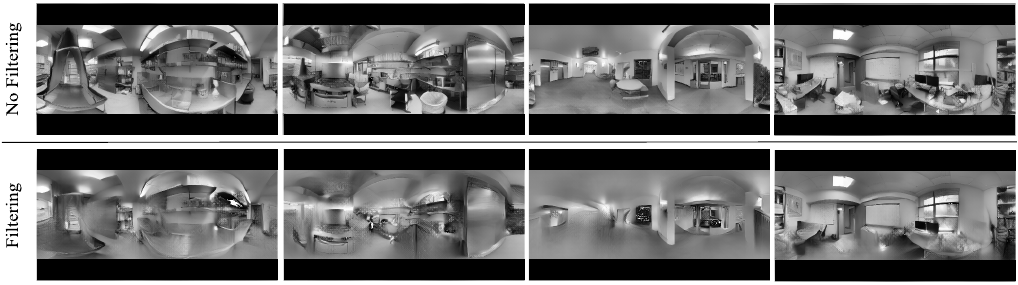}
    \caption{Deletion of objects in feature inversion attacks after line-based filtering.}
   \label{fig:qualitative_additional}
\vspace{-1em}
\end{figure*}
\section{Baseline Details}
In this section, we describe the details for implementing the baselines compared against LDL.
We implement PICCOLO~\cite{piccolo} and CPO~\cite{cpo} from the publicly available codebase.
Below we retain our description on the Structure-based, Chamfer-based, and Line Transformer-based approaches.

\paragraph{Structure-Based Approach}
As explained in Section 4, structured-based approach first finds promising candidate poses using  robust image retrieval and then refines poses using PnP-RANSAC from feature matches.
For image retrieval we use NetVLAD~\cite{netvlad}, which is a widely used image retrieval method that outputs a global feature vector for each image.
To deploy NetVLAD in our setup, we first render $N_t \times N_r$ synthetic views from the point cloud.
Here we use $N_t=100$ candidate translations and $N_r=216$ candidate rotations uniformly sampled from $SO(3)$.
Then, we extract the global features for each synthetic view and the query image, and choose the top $K=20$ synthetic views whose feature vectors are closest to the query image.
As the final step, we perform feature matching~\cite{sarlin2020superglue} from each selected synthetic view against the query image, and choose the final view with the most matches.
To ensure fair comparison, we undistort the selected view and the query panorama into cubemaps and separately perform feature matching for each pair of faces.
The matches are then aggregated to perform refinement via PnP-RANSAC~\cite{ransac}.

\paragraph{Chamfer Distance-Based Approach}
Inspired from Micusik et al.~\cite{line_chamfer}, Chamfer distance-based approach first selects poses that best align 3D lines against lines in the query image, where the Chamfer distance is used to evaluate the potential matchings.
The selected poses are then refined with PnP-RANSAC, similar to the structure-based approach.
To elaborate, we find the top $K=20$ poses from an initial pool of $N_t \times N_r$ poses, where the poses are ranked by measuring the Chamfer distance between the projected line segments in 3D and those in the query image.
We set $N_t$ and $N_r$ identical to LDL and use the principal directions for deducing a set of candidate rotations.
As the final step, we render views at the selected $K$ poses and perform feature matching against the query image for refinement via PnP-RANSAC.

\paragraph{Line Transformer-Based Approach}
Based on Yoon et al.~\cite{line_transformer}, Line Transformer-based approach finds candidate poses attaining the most line matches with the query image, and refines poses using PnP-RANSAC.
For establishing line matches, we first render $N_t \times N_r$ synthetic views from the point cloud where we set $N_t=100$ and $N_r=216$.
Then, the top $K_1=100$ poses are selected whose NetVLAD~\cite{netvlad} features are closest to the query image.
This intermediate step is necessary as the line transformer features are computationally expensive and thus could not be naively evaluated for all $N_t \times N_r$ views.
For each synthetic view from the selected poses, we extract line Transformer embeddings and establish matchings with the query image.
Similar to the structure-based baseline, we convert panoramas to cubemaps during the line matching process.
Finally, we select the top $K_2=20$ poses that have the most line matches, and refine them via PnP-RANSAC.

\section{Details on Experimental Setup}

In this section, we provide additional details for experiments presented in Section 4 and Section~\ref{sec:ablation}.

\begin{figure}[t]
  \centering
    \includegraphics[width=\linewidth]{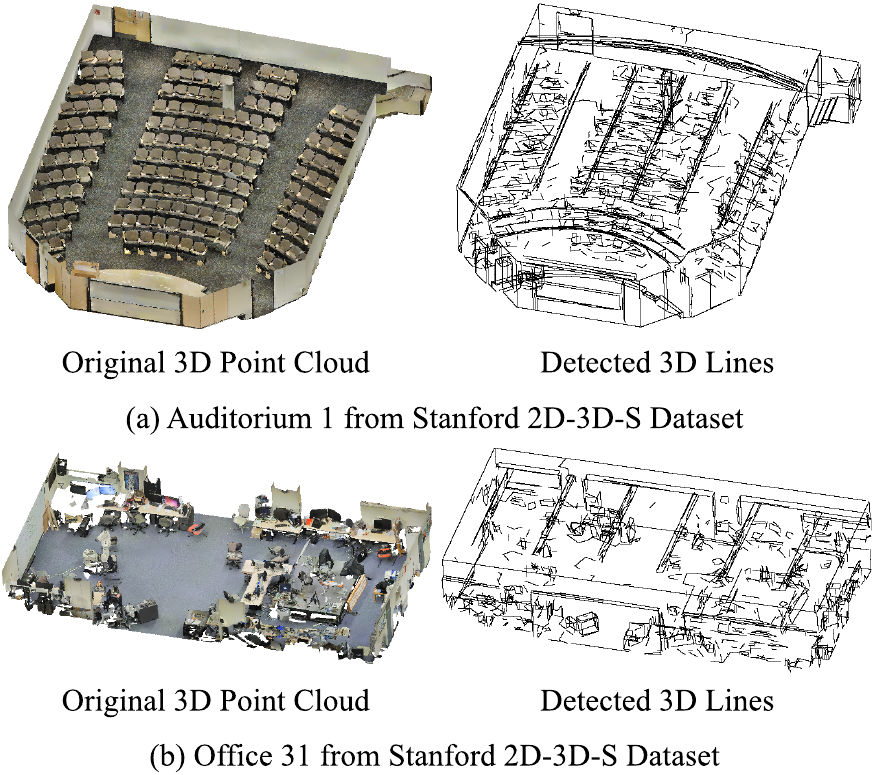}
   \caption{Visualization of the 3D line segments used for LDL. While the line segment extraction algorithm from Xiaohu et al.~\cite{3d_lineseg} can reliably extract the wireframe-like structure from the original 3D scan, the line segments are still quite noisy. Note that we have cropped the ceilings of the original point cloud for better visualization.}
   \label{fig:line_map}
\vspace{-1em}
\end{figure}
\paragraph{Illumination Robustness Evaluation}
To evaluate the robustness of LDL against illumination shifts, we apply synthetic color variations to images in Room 3 from OmniScenes~\cite{piccolo}.
We consider three synthetic color variations, where qualitative examples are shown in Figure 4: average intensity, gamma, and white balance change.
For average intensity change we lower each pixel intensity by 25\%.
For gamma change, we set the image gamma to 0.2.
For white balance change, we apply the following transformation matrix to the raw RGB color values: $\begin{pmatrix}
0.7 & 0 & 0\\
0 & 0.9 & 0\\
0 & 0 & 0.8
\end{pmatrix}$.

\paragraph{Candidate Pose Search Evaluation}
We compare LDL against NetVLAD~\cite{netvlad} for candidate pose search using the Extreme split from OmniScenes.
The recall curves in Figure 5 are obtained by measuring the localization performance of both methods prior to pose refinement.
As mentioned in Section 4.2, we use the identical set of translations with $N_t=50$ for both methods and associate a large number of candidate rotations $N_r=216$ for NetVLAD to ensure fair comparison.
Such measures are taken for rotations since LDL estimates rotations using combinatorial matchings of principal directions, which makes the number of candidate rotations to vary for each query image.
We empircally find that less than 30 candidate rotations remain after discarding infeasible rotations, and thus setting $N_r=216$ for NetVLAD would provide enough evidence to achieve competitive performance against LDL.

\paragraph{Feature Inversion Network for Privacy Evaluation}
To evaluate the privacy protection of LDL against feature inversion attacks, we train a fully-convolutional neural network $F_\Theta(\cdot)$ that takes a sparse feature map $D \in \mathbb{R}^{H \times W \times C}$ as input and produces image reconstructions.
The feature map stores local feature descriptors $\mathbf{f} \in \mathbb{R}^C$ at keypoint locations $(i_\text{kpt}, j_\text{kpt})$, namely $D(i_\text{kpt}, j_\text{kpt}) = \mathbf{f}$, and zero values for other regions.
For the inversion network, we use a similar U-Net architecture as in Ng et al.~\cite{ninjadesc} where the only difference is in the input channel dimension that we set as 256 instead of 128 to match the SuperPoint~\cite{superpoint} descriptor dimensions.
Then for training, we use the entire Matterport3D~\cite{matterport} dataset where we use the first $90\%$ of the 9581 panorama images for training and the rest for validation.
We follow the training procedure of Ng et al.~\cite{ninjadesc}  and use the perceptual loss and mean absolute error (MAE) loss, where we employ Adam~\cite{adam} with a learning rate of $1e{-}4$ for optimization.
In our experiments, we use the trained network to reconstruct panoramas from the local feature descriptors, where we shared the reconstruction results along with the image error metrics in Section 4 and Section~\ref{sec:ablation}.
To elaborate, during evaluation we first extract local features for each query image in the Stanford 2D-3D-S dataset~\cite{stanford2d3d} and run feature inversion, where the results are then compared against the original panorama image.

\paragraph{3D Line Maps for Localization}
In Figure~\ref{fig:line_map}, we show visualizations of 3D lines used as input to LDL.
Despite the reliabilty of the 3D line extraction algorithm of Xiaohu et al.~\cite{3d_lineseg}, the lines are still quite noisy.
To cope with the noisy detections, LDL employs a length-based filtering scheme to only keep long, salient lines and resorts to matching the \textit{distribution} of lines using line distance functions instead of trying to establish direct one-to-one matchings as in previous works~\cite{line_chamfer,line_transformer}.

{\small
\bibliographystyle{ieee_fullname}
\bibliography{main}
}

\end{document}